\documentclass{article}

\usepackage[main,final]{neurips_2026}

\usepackage[utf8]{inputenc} %
\usepackage[T1]{fontenc}    %
\usepackage{url}            %
\usepackage{booktabs}       %
\usepackage{amsfonts}       %
\usepackage{amsmath}        %
\usepackage{amssymb}        %
\usepackage{amsthm}         %
\newtheorem{lemma}{Lemma}[section]
\newtheorem{proposition}[lemma]{Proposition}
\newtheorem{theorem}[lemma]{Theorem}
\newtheorem{corollary}[lemma]{Corollary}
\usepackage{nicefrac}       %
\usepackage{microtype}      %
\usepackage[table]{xcolor}  %
\usepackage{colortbl}       %
\usepackage{graphicx}       %
\usepackage{placeins}      %
\usepackage{wrapfig}        %
\usepackage{algorithm}      %
\usepackage{algpseudocode}  %
\usepackage{titletoc}       %
\usepackage{enumitem}       %
\usepackage[hypertexnames=false]{hyperref} %
\ifdefined\pdfminorversion
\fi

\definecolor{secondcyan}{HTML}{1B7CB6}  %
\definecolor{enginerow}{HTML}{E8F2FF}   %
\definecolor{engineorange}{HTML}{DD8452}  %
\definecolor{engineblue}{HTML}{4C72B0}    %
\definecolor{enginepurple}{HTML}{9B30A0}  %
\definecolor{enginegreen}{HTML}{55A868}   %
\newcommand{\best}[1]{\textbf{#1}}
\newcommand{\snd}[1]{\textcolor{secondcyan}{\textbf{#1}}}
\newcommand{\spm}[1]{{\scriptsize\,$\pm$\,#1}}
\definecolor{gaingreen}{HTML}{22C55E}  %
\newcommand{\gain}[1]{\textcolor{gaingreen}{+#1}}
\newcommand{\Engine}{\textsc{Engine}}

\title{ENGINE: Endogenous Variational MultiScale Optimization for Zeroth-Order LLM Fine-Tuning}

\author{%
  Zhuoli Ouyang$^{1,2}$\thanks{Work done as an intern at Westlake University.}\quad
  Changxi Chi$^{1,3}$\quad
  Siyuan Li$^{1,3}$\quad
  Tailin Wu$^{1}$\thanks{Corresponding author.} \\
  $^{1}$Department of Artificial Intelligence, School of Engineering, Westlake University \\
  $^{2}$Department of Electronic and Electrical Engineering, \\
  Southern University of Science and Technology \\
  $^{3}$Zhejiang University \\
  \texttt{ouyangzl2023@mail.sustech.edu.cn}\quad\texttt{wutailin@westlake.edu.cn}
}

\begin{document}

\setcounter{footnote}{1}
\maketitle
\setcounter{footnote}{0}

\begin{abstract}
Zeroth-order (ZO) optimization enables memory-efficient large language model (LLM) fine-tuning, and recent methods incorporate curvature-aware strategies. However, in the high-dimensional parameter space of LLMs, these methods inherently suffer from exploding gradient variance and prohibitive costs. Conversely, confining updates to a low-dimensional subspace reduces variance but discards critical high-dimensional information, degrading performance. This creates a fundamental dilemma: full-space optimization is informative but has high variance, while subspace optimization has low variance but loses information. To break this trade-off, we propose \textbf{\Engine} (\textbf{En}do\textbf{g}enous Var\textbf{i}atio\textbf{n}al MultiScal\textbf{e}). Inspired by computational mechanics, we adapt static Variational MultiScale (VMS) theory to dynamic optimization, treating low- and high-dimensional spaces as resolved coarse and unresolved fine scales. Through a multigrid-style V-cycle, \Engine\ alternates between spaces, dynamically probing fine-grained information in the high-dimensional space and injecting it to correct the efficient low-dimensional sampling process. Under appropriate assumptions, statistical analysis characterizes variance reduction and correction of curvature-induced bias, and convergence analysis establishes a stationarity bound for the corrected coarse updates. Evaluations on standard LLM benchmarks show that \Engine\ achieves higher accuracy and faster convergence than existing ZO methods, achieving up to a $\mathbf{2.61\times}$ \textbf{speed-up in forward passes} with zero additional peak memory overhead. Code is available at \textcolor{secondcyan}{\url{https://github.com/AI4Science-WestlakeU/ENGINE}}.

\end{abstract}

\section{Introduction}
\label{sec:intro}
\begin{wraptable}[10]{r}{0.46\textwidth}
  \centering
  \small
  \caption{Trade-offs between full-space and subspace ZO optimization.
           $\checkmark$ denotes a relative strength, and $\times$ a weakness.}
  \label{tab:intro-comparison}
  \renewcommand{\arraystretch}{1.0}
\resizebox{1.0\linewidth}{!}{
  \begin{tabular}{lcccc}
    \toprule
    \textbf{Space} & \textbf{Effic.} & \textbf{Var.} & \textbf{Perf.} & \textbf{Curv.} \\
    \midrule
    Full space   & \textcolor{red}{$\times$}      & \textcolor{red}{$\times$}      & \textcolor{green!60!black}{$\checkmark$}  & \textcolor{green!60!black}{$\checkmark$} \\
    Subspace     & \textcolor{green!60!black}{$\checkmark$}  & \textcolor{green!60!black}{$\checkmark$}  & \textcolor{red}{$\times$}      & \textcolor{red}{$\times$}     \\
    \textbf{\textsc{\Engine} (Ours)}& \textcolor{green!60!black}{$\checkmark$}  & \textcolor{green!60!black}{$\checkmark$}  & \textcolor{green!60!black}{$\checkmark$}  & \textcolor{green!60!black}{$\checkmark$} \\
    \bottomrule
  \end{tabular}
  }
\end{wraptable}

Large language model fine-tuning relies on first-order (FO) optimizers~\citep{loshchilov2017fixing}.
However, FO training requires storing activations, gradients, and optimizer states, making it memory-intensive.
Zeroth-order (ZO) optimization~\citep{malladi2023fine} is therefore an appealing alternative: it uses only
forward passes and reduces memory to near inference level.
Still, in the full LLM parameter space, ZO methods suffer from high variance and high query
cost~\citep{nesterov2017random}.
LoRA~\citep{hu2021lora} offers a lower-dimensional alternative by restricting updates to a low-rank subspace,
improving efficiency and usually reducing variance.
But this also discards fine-grained curvature information from the full space, which can slow convergence and degrade performance.

This leads to a central dilemma in zeroth-order optimization (Table~\ref{tab:intro-comparison}):
\textbf{full-space optimization is informative but high-variance, while subspace optimization is low-variance but blind.}
Recent curvature-aware ZO methods, such as LOREN~\citep{seung2026low}, reduce estimator variance through
anisotropic perturbations~\citep{kool2019buy}.
Even so, existing methods usually stay in one space only: either the full parameter space or a low-rank subspace
such as LoRA.
They therefore gain only one side of the trade-off.
Full-space methods retain richer information but are statistically costly, while subspace methods are more
efficient but inevitably lose part of the curvature structure that is present in the full space.
This motivates a simple question:
\textbf{can we keep the efficiency of low-dimensional sampling while still using curvature information from the
full space?}
We approach this question through Variational MultiScale
(VMS) theory~\citep{hughes1995multiscale,hughes1998variational} from computational mechanics.
VMS separates a problem into a resolved coarse scale and an unresolved fine scale, then uses a correction
term~\citep{hughes2000large} to account for the effect of the unresolved part.
This viewpoint suggests a natural way to think about low-dimensional and high-dimensional spaces in zeroth-order
optimization.

Based on this idea, we propose \textbf{\Engine}
(\textbf{En}do\textbf{g}enous Var\textbf{i}atio\textbf{n}al MultiScal\textbf{e}).
We treat the low-dimensional LoRA parameterization as the resolved coarse scale and incorporate unresolved full-space curvature through a fine-scale correction.
Unlike classical VMS, the optimization operator evolves with the optimization state, so the interaction between
the two spaces must be updated dynamically.
\Engine\ addresses this with an endogenous V-cycle that alternates between the two spaces during training.
Through a closed-form subgrid correction, it uses fine-scale information from the high-dimensional space to
correct the efficient low-dimensional update.

\begin{figure}[t]
\centering
\includegraphics[width=\linewidth]{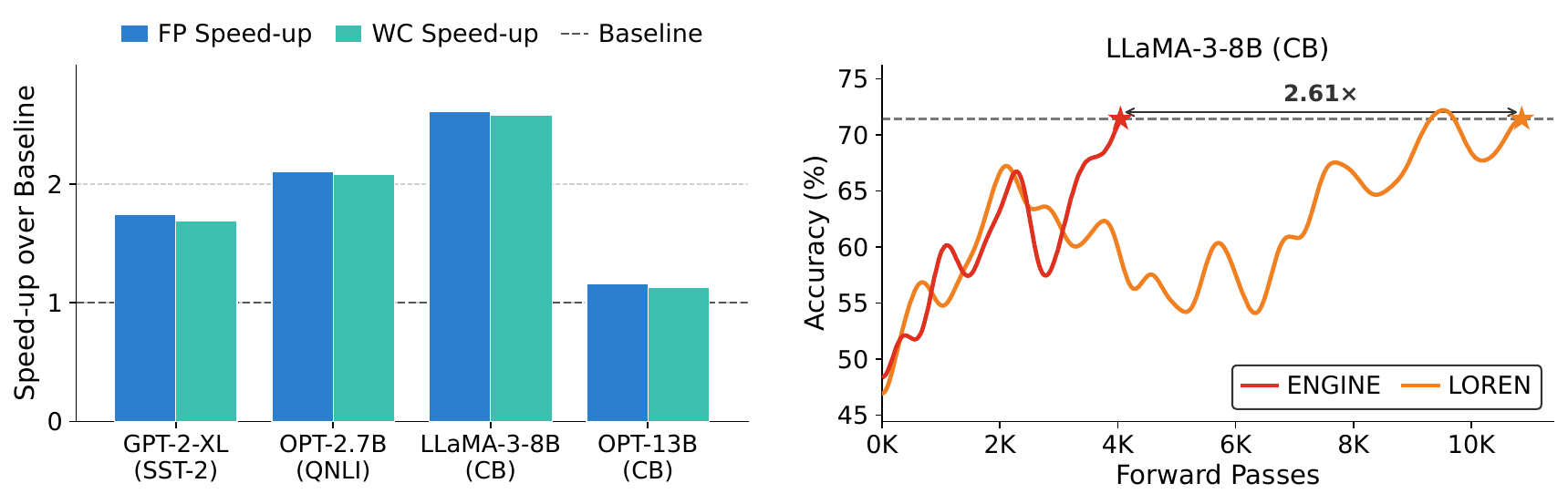}
\caption{%
  \textbf{Left:} Forward-pass and wall-clock speed-up of \Engine\ over LOREN to reach LOREN's best accuracy,
  across four representative (model, task) settings.
  \textbf{Right:} Validation accuracy convergence on LLaMA-3-8B (CB).}
\label{fig:speedup-convergence}
\end{figure}

Experiments on standard LLM optimization benchmarks show that \Engine\ consistently improves over state-of-the-art zeroth-order baselines, including LOREN.
It achieves better downstream accuracy, faster convergence, and up to a \textbf{$2.61\times$ speed-up in forward passes} (Figure~\ref{fig:speedup-convergence}),
while keeping peak memory usage unchanged.

\begin{figure}[t]
  \centering
  \includegraphics[width=\linewidth]{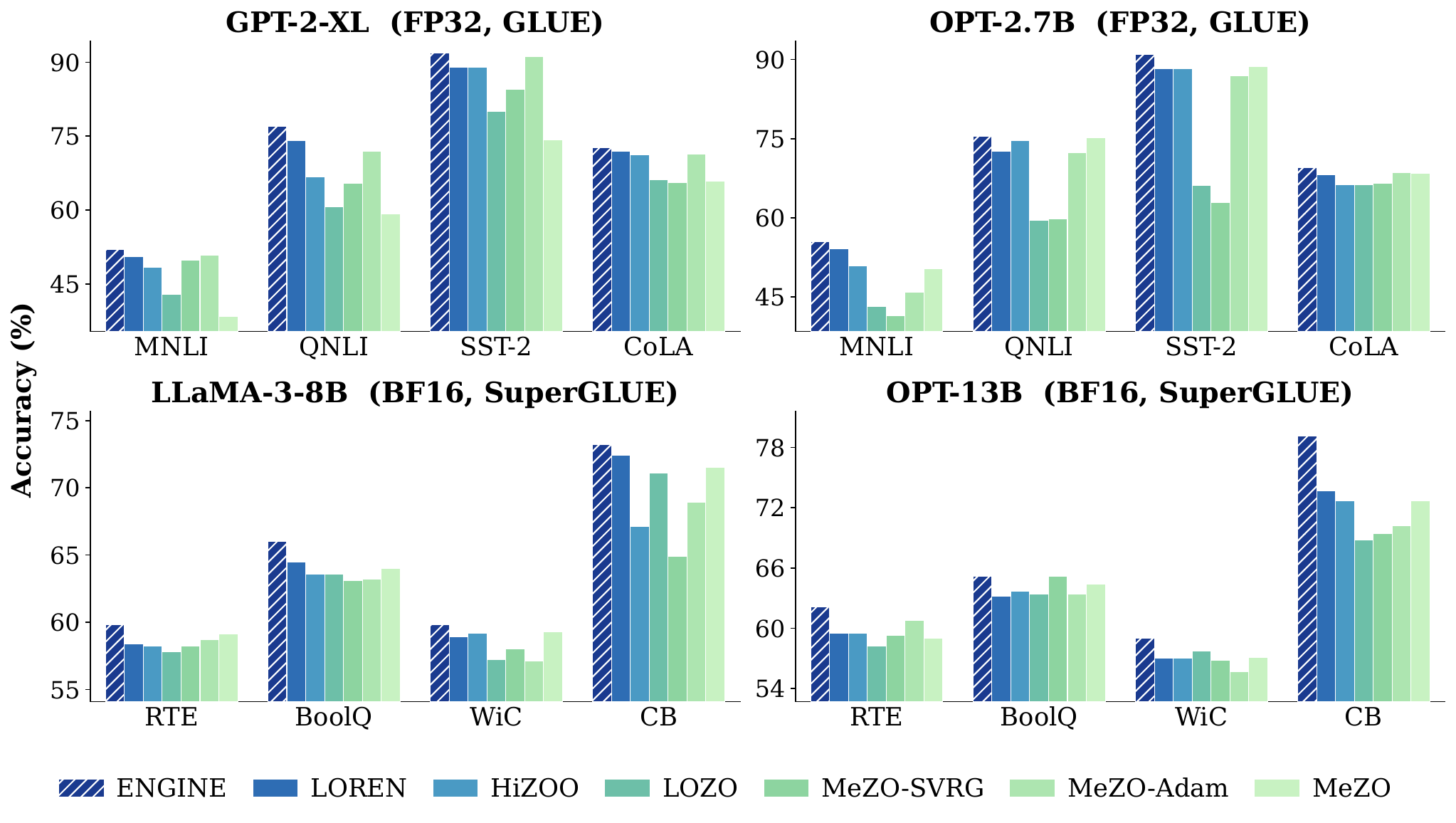}
  \caption{Test accuracy on all 16 (model, task) combinations.
    \textbf{Top row}: GPT-2-XL (GLUE, FP32) and OPT-2.7B (GLUE, FP32).
    \textbf{Bottom row}: LLaMA-3-8B (SuperGLUE, BF16) and OPT-13B (SuperGLUE, BF16).
    \Engine\ is shown in \textbf{blue hatched}.
    Full numerical values are in Appendix~\ref{sec:app-main-tables}.}
  \label{fig:bars-main}
\end{figure}

Our contributions are summarised as follows.
\begin{itemize}
  \item \textbf{A cross-disciplinary perspective.}
        We introduce the idea of Variational MultiScale (VMS) from computational mechanics
        into zeroth-order LLM fine-tuning.
        This perspective clarifies why unresolved fine-scale curvature matters and provides a systematic way
        to describe the interaction between low-dimensional and high-dimensional optimization spaces.

  \item \textbf{The framework of \Engine.}
        We propose \Engine, an endogenous Variational MultiScale framework for zeroth-order optimization that
        combines low-rank sampling with a closed-form subgrid correction, thereby reintroducing unresolved
        full-space curvature into the coarse update.
        Under appropriate assumptions, we rigorously analyze the bias and variance of the corrected estimator.
        We also establish a stationarity bound for coarse updates.

  \item \textbf{Consistent improvements.}
        Extensive experiments across diverse LLM fine-tuning settings demonstrate that \Engine\ improves both
        downstream performance and convergence speed over state-of-the-art zeroth-order baselines (Figure~\ref{fig:bars-main}).
        These gains come with up to a $2.61\times$ speed-up in forward passes and no additional peak memory
        overhead.
\end{itemize}

\section{Preliminaries \& Problem Setup}
\label{sec:prelim}
\noindent To address the single-scale limitation of ZO methods discussed in Sec.~\ref{sec:intro}, \Engine\ couples a low-dimensional LoRA scale with the high-dimensional full parameter space via Variational MultiScale (VMS). Sec.~\ref{sec:prelim-zo} first introduces the curvature-aware ZO framework adopted from \textsc{Loren}, after which Sec.~\ref{sec:prelim-vms} recalls the Variational MultiScale (VMS) viewpoint underlying our method.

\subsection{Zeroth-Order Setting \& Notation Setup}
\label{sec:prelim-zo}

\textbf{Notation.} We denote vectors by lowercase bold and matrices by uppercase bold, where $\otimes$ is the Kronecker product and $\|\cdot\|$ the Euclidean norm. For a weight matrix $\mathbf{X}\in\mathbb{R}^{m\times n}$, let $\mathbf{x}:=\mathrm{vec}(\mathbf{X})\in\mathbb{R}^{d}$ with $d=mn$ (multi-layer models are handled block-wise). We minimize $f(\mathbf{x}):=\mathbb{E}_{\xi\sim\mathcal{P}_{\mathrm{data}}}[\ell(\mathbf{x},\xi)]$ using only function-value queries, where $\mathcal{P}_{\mathrm{data}}$ is the data distribution.

\textbf{Rank-1 Kronecker curvature.} Full Hessian storage costs $O(m^2n^2)$ memory. K-FAC~\citep{martens2015optimizing} and Shampoo~\citep{gupta2018shampoo} exploit matrix structure to reduce preconditioning costs. Following \textsc{Loren}, we use the compact vector parameterization:
\begin{equation}\label{eq:prelim-rank1}
\widetilde{\mathbf{H}}\;=\;\mathbf{I}_{m}\otimes\bigl(\rho\mathbf{I}_{n}+\mathbf{v}\mathbf{v}^{\top}\bigr),\qquad \mathbf{v}\in\mathbb{R}^{n},\ \rho>0,
\end{equation}
where $\mathbf{v}\mathbf{v}^{\top}$ captures anisotropy shared across blocks, and damping ensures positive definiteness. Storing $\mathbf{v}$ costs $O(n)$. The inverse and inverse square root admit closed forms with $O(mn)$ application cost, avoiding dense matrices. Section~\ref{sec:method} extends this approximation to multiple scales.

\textbf{Curvature as covariance.} For $\mathbf{u}\sim\mathcal{N}(\mathbf{0},\mathbf{I}_{d})$ and smoothing radius $\epsilon>0$, the symmetric finite-difference estimator~\citep{spall1992multivariate} is $\widehat{\mathbf g}_\epsilon(\mathbf{x},\mathbf{u})=\tfrac{f(\mathbf{x}+\epsilon\mathbf{u})-f(\mathbf{x}-\epsilon\mathbf{u})}{2\epsilon}\mathbf{u}$. Its expectation is the gradient of the Gaussian-smoothed loss. Generalized Gaussian smoothing extends this to anisotropic perturbations $\mathbf{u}\sim\mathcal{N}(\mathbf{0},\boldsymbol{\Sigma})$, and setting $\boldsymbol{\Sigma}=\widetilde{\mathbf{H}}^{-1}$ yields the NES identity~\citep{wierstra2014natural} $\widetilde{\mathbf{H}}^{-1}\nabla f(\mathbf{x})=\mathbb{E}_{\tilde{\mathbf{u}}\sim\mathcal{N}(\mathbf{0},\widetilde{\mathbf{H}}^{-1})}\!\bigl[\langle\nabla f(\mathbf{x}),\tilde{\mathbf{u}}\rangle\,\tilde{\mathbf{u}}\bigr]$, so preconditioning by $\widetilde{\mathbf{H}}^{-1}$ corresponds to larger ZO perturbations along directions of lower estimated curvature.

\subsection{Variational MultiScale Method}
\label{sec:prelim-vms}

The Variational MultiScale (VMS) method decomposes problems into resolved (coarse) and unresolved (fine) scales. Let $\Omega\subset\mathbb{R}^d$ be a bounded domain, $\mathcal{V}$ a function space over $\Omega$ with dual $\mathcal{V}^*$, and $\mathcal{L}:\mathcal{V}\to\mathcal{V}^*$ a linear operator, where the symmetric bilinear form $a(w,u):=(w,\mathcal{L}u)$ uses the $L^2(\Omega)$ inner product $(w,u):=\int_{\Omega}w(x)\,u(x)\,\mathrm{d}x$. The weak problem seeks $u\in\mathcal{V}$ such that $a(w,u)=(w,f)$ for all test functions $w\in\mathcal{V}$. VMS decomposes $\mathcal{V}=\bar{\mathcal{V}}\oplus\mathcal{V}'$ into coarse $\bar{\mathcal{V}}$ and fine $\mathcal{V}'$ subspaces, so that $u=\bar u+u'$ and $w=\bar w+w'$ with $\bar u,\bar w\in\bar{\mathcal{V}}$ and $u',w'\in\mathcal{V}'$. The fine scale is driven by the coarse residual $\mathcal{L}u'=-(\mathcal{L}\bar u-f)$. Following the sign convention of \citet{hughes1995multiscale}, the fine-scale solution operator $M:\mathcal{V}^*\to\mathcal{V}'$ includes the negative sign, so $u'=M(\mathcal{L}\bar u-f)$. Substitution yields the closed coarse equation:
\begin{equation}\label{eq:prelim-vms-closed}
a(\bar w,\bar u)\;+\;\bigl(\mathcal{L}^{*}\bar w,\,M(\mathcal{L}\bar u-f)\bigr)\;=\;(\bar w,f),
\end{equation}
where $\mathcal{L}^{*}:\mathcal{V}\to\mathcal{V}^*$ is the adjoint of $\mathcal{L}$, and the second term in~\eqref{eq:prelim-vms-closed} is the \emph{subgrid correction} encoding unresolved-scale feedback. In our setting, $\bar u$, $u'$, and $\mathcal{L}$ correspond to the LoRA-space update, unresolved full-space curvature, and endogenous Hessian operator $\widetilde{\mathbf{H}}$, respectively. The subgrid correction corresponds to the VMS-EO fine-scale curvature injection developed in Sec.~\ref{sec:method}.

\section{Method}
\label{sec:method}
\begin{figure}[t]
  \centering
  \includegraphics[width=\linewidth]{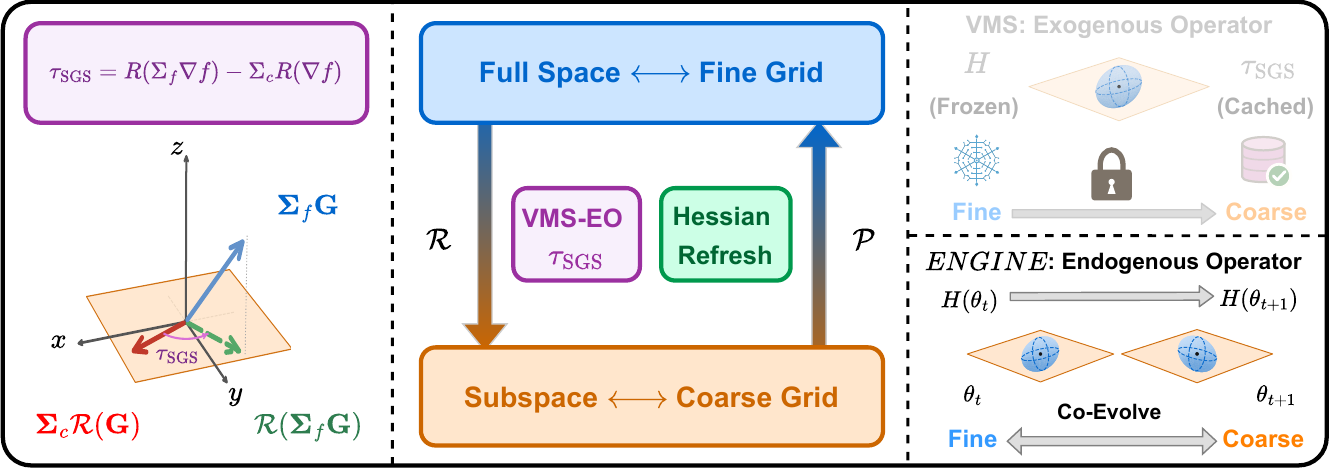}
  \caption{\textbf{ENGINE overview.}
           \emph{Left}: the gap between the restricted fine direction $\mathcal{R}(\boldsymbol{\Sigma}_f\mathbf{G})$ and the coarse direction $\boldsymbol{\Sigma}_c\mathcal{R}(\mathbf{G})$ defines the subgrid correction $\boldsymbol{\tau}_{\mathrm{sgs}}$.
           \emph{Middle}: V-cycle alternates Coarse ($\boldsymbol{\phi}$) and Fine ($\mathbf{W}$) phases, with every $S$ steps a fine probe refreshing $\mathbf{v}_f\!\leftarrow\!\mathbf{v}_f\!-\!\nu g(\mathbf{v}_f)$ and $\lambda$ (Hessian refresh), while VMS-EO ($\boldsymbol{\tau}_{\mathrm{sgs}}$) injects the subgrid correction into each coarse step.
           \emph{Right}: classical VMS uses a fixed operator $\mathcal{L}$, while \Engine\ uses the dynamic Hessian $\mathbf{H}(\boldsymbol{\theta})$.}
  \label{fig:concept}
\end{figure}

Building on the rank-1 curvature primitive $(\widetilde{\mathbf{H}},\mathbf{v})$ from Eq.~\eqref{eq:prelim-rank1} and the VMS formulation coupling coarse and fine scales from Eq.~\eqref{eq:prelim-vms-closed}, this section develops and analyzes \Engine\ (Figure~\ref{fig:concept}). \S\ref{sec:method-variance} contrasts the \emph{rigid} linear operator $\mathcal{L}$ of classical VMS with the \emph{dynamic} Hessian $\mathbf{H}(\boldsymbol{\theta})$ of optimization, motivating why classical VMS cannot be ported verbatim. \S\ref{sec:method-design} derives the central correction $\boldsymbol{\tau}_{\mathrm{sgs}}$ and its rank-1 LoRA-branch surrogate. \S\ref{sec:method-vcycle} embeds VMS-EO into a V-cycle so that the endogenous state $\mathbf{v}_f$ is refreshed across phases. \S\ref{sec:method-statistical} then quantifies how coarse sampling and exact subgrid correction affect variance and bias relative to a common target.

The multiscale state is carried by three rank-1 curvature vectors: the full-space $\mathbf{v}_f\in\mathbb{R}^{n}$ and the two LoRA-branch curvatures $\mathbf{v}_{\!\mathbf{B}}\in\mathbb{R}^{r}$, $\mathbf{v}_{\!\mathbf{A}}\in\mathbb{R}^{n}$. We instantiate the resolved coarse scale by a rank-$r$ LoRA factorization $\Delta\mathbf{W}=\mathbf{B}\mathbf{A}$ with $\mathbf{B}\in\mathbb{R}^{m\times r}$, $\mathbf{A}\in\mathbb{R}^{r\times n}$ and $r\ll\min(m,n)$, so the coarse parameter $\boldsymbol{\phi}=(\mathbf{B},\mathbf{A})$ has dimension $d_c=r(m+n)$, much smaller than $d_f=mn$. Fine-scale information is obtained from the full parameter space and used to correct curvature effects not captured by the coarse model.

\subsection{Endogenous Operator: Why Classical VMS Fails?}
\label{sec:method-variance}

Classical VMS starts from a linear problem $\mathcal{L}u=f$ with operator $\mathcal{L}$ fixed by the underlying physics, which is what allows the unresolved component $u'$ to be eliminated analytically in closed form (Figure~\ref{fig:concept} Right, top). In optimization the analogous linear equation is the Newton step $\mathbf{H}(\boldsymbol{\theta})\Delta\boldsymbol{\theta}=-\nabla f(\boldsymbol{\theta})$, giving the correspondence $(\mathcal{L},u,f)\leftrightarrow(\mathbf{H},\Delta\boldsymbol{\theta},-\nabla f)$. Unlike $\mathcal{L}$, however, $\mathbf{H}(\boldsymbol{\theta})$ co-evolves with the iterate (Figure~\ref{fig:concept} Right, bottom), so the closure cannot be derived once and for all but must be refreshed periodically by the V-cycle of \S\ref{sec:method-vcycle}. We call this state-dependent operator an endogenous operator and the resulting subgrid correction VMS-EO.

Coarse increments are represented by changes in the LoRA factors. The \emph{prolongation} operator $\mathcal{P}:(\Delta\mathbf{B},\Delta\mathbf{A})\longmapsto\mathbf{B}\Delta\mathbf{A}+\Delta\mathbf{B}\mathbf{A}\in\mathbb{R}^{m\times n}$ lifts a coarse update to the full space (the differential of the LoRA product), and the \emph{restriction} operator $\mathcal{R}=(\mathcal{R}_{\!\mathbf{B}},\mathcal{R}_{\!\mathbf{A}})$ with $\mathcal{R}_{\!\mathbf{B}}(\mathbf{X})=\mathbf{X}\mathbf{A}^{\top}\in\mathbb{R}^{m\times r}$ and $\mathcal{R}_{\!\mathbf{A}}(\mathbf{X})=\mathbf{B}^{\top}\mathbf{X}\in\mathbb{R}^{r\times n}$ maps a full-space matrix $\mathbf{X}$ to the two LoRA gradient blocks. The two operators are formal adjoints under the Frobenius inner product, $\mathcal{R}=\mathcal{P}^{*}$, i.e., $\langle\mathcal{P}(\Delta\mathbf{B},\Delta\mathbf{A}),\mathbf{X}\rangle_{F}=\langle(\Delta\mathbf{B},\Delta\mathbf{A}),\mathcal{R}(\mathbf{X})\rangle_{F}$ for all $\mathbf{X}\in\mathbb{R}^{m\times n}$.

At LLM scale, $\mathbf{H}\in\mathbb{R}^{d_f\times d_f}$ cannot be formed, so \Engine\ inherits LOREN's rank-1 Kronecker $\widetilde{\mathbf{H}}=\mathbf{I}_m\otimes(\rho\mathbf{I}_n+\mathbf{v}\mathbf{v}^{\top})$ as the practical operator with preconditioner $\boldsymbol{\Sigma}=\widetilde{\mathbf{H}}^{-1}$. Given parameter $\mathbf{x}$ and curvature $\mathbf{v}$, LOREN draws $K$ anisotropic perturbations $z_k=\mathbf{x}+\varepsilon\boldsymbol{\Sigma}^{1/2}u_k$ where $u_k\sim\mathcal{N}(\mathbf{0},\mathbf{I})$ to simultaneously form RLOO estimators for both parameter and curvature gradients:
\begin{equation}\label{eq:method-rloo}
g(\mathbf{x})\;=\;\frac{1}{\varepsilon(K-1)}\sum_{k=1}^{K}(f_k-\bar f)\,\boldsymbol{\Sigma}^{1/2}u_k,\qquad g(\mathbf{v})\;=\;\frac{1}{K-1}\sum_{k=1}^{K}(f_k-\bar f)\,\nabla_{\mathbf{v}}\log p_{\boldsymbol{\theta}}(z_k),
\end{equation}
where $f_k:=f(z_k)$, $\bar f:=\tfrac{1}{K}\sum_j f_j$, and the closed form of the score function $\nabla_{\mathbf{v}}\log p_{\boldsymbol{\theta}}(z)$ is given in LOREN Eq.~(10). The two estimators \emph{share the same batch} $\{u_k,f_k\}$, so each step costs only $K$ forward passes. \Engine\ samples in LoRA coordinates with~\eqref{eq:method-rloo} to avoid $d_f$-scale variance, while injecting full-space curvature feedback as a subgrid correction back into the coarse phase.

\subsection{VMS-EO Subgrid Correction}
\label{sec:method-design}

To construct the correction in Figure~\ref{fig:concept} (Left), we first identify where coarse and fine directions differ. Lemma~\ref{lem:chain-rule} establishes exact restriction of first-order gradients. Proposition~\ref{prop:subgrid-residual} then gives the curvature mismatch introduced by preconditioning. This residual supplies the corrected LoRA updates in Eqs.~\eqref{eq:method-vms-B} and \eqref{eq:method-vms-A}.

\paragraph{Closed-Form Curvature Correction.}
\label{sec:method-vms-eo}

Let $\mathbf{G}\in\mathbb{R}^{m\times n}$ denote a full-space gradient-like matrix. Deterministically $\mathbf{G}=\nabla_{\!\mathbf{W}}f$, while in the ZO implementation it is replaced by the RLOO estimator.

\begin{lemma}[chain rule]\label{lem:chain-rule}
LoRA restriction commutes with first-order differentiation: $\mathcal{R}(\nabla_{\!\mathbf{W}}f)=\nabla_{(\mathbf{B},\mathbf{A})}f$. Proof in Appendix~\ref{sec:app-lemma-51}.
\end{lemma}

Lemma~\ref{lem:chain-rule} shows that restricting the exact fine-space gradient gives the coarse-space gradient. Curvature preconditioning introduces a mismatch between the two directions. Let $\boldsymbol{\Sigma}_f,\boldsymbol{\Sigma}_{\!\mathbf{B}},\boldsymbol{\Sigma}_{\!\mathbf{A}}$ be the rank-1 Kronecker covariances built from $\mathbf{v}_f,\mathbf{v}_{\!\mathbf{B}},\mathbf{v}_{\!\mathbf{A}}$ (Sherman-Morrison forms in Appendix~\ref{sec:app-thm-52}), and $\boldsymbol{\Sigma}_c:=\mathrm{diag}(\boldsymbol{\Sigma}_{\!\mathbf{B}},\boldsymbol{\Sigma}_{\!\mathbf{A}})$. In general $\mathcal{R}(\boldsymbol{\Sigma}_f\mathbf{G})\neq\boldsymbol{\Sigma}_c\mathcal{R}(\mathbf{G})$, so the two preconditioned directions can differ in both magnitude and orientation. This gap defines the \emph{subgrid residual}:
\begin{equation}\label{eq:method-tau-sgs}
\boldsymbol{\tau}_{\mathrm{sgs}}\;:=\;\mathcal{R}(\boldsymbol{\Sigma}_f\mathbf{G})-\boldsymbol{\Sigma}_c\mathcal{R}(\mathbf{G}),
\end{equation}
with branch components $\boldsymbol{\tau}_{\mathrm{sgs},\mathbf{B}}\in\mathbb{R}^{m\times r}$ and $\boldsymbol{\tau}_{\mathrm{sgs},\mathbf{A}}\in\mathbb{R}^{r\times n}$. The following result expresses each branch residual through the rank-1 curvature vectors.

\begin{proposition}[closed-form subgrid residual]\label{prop:subgrid-residual}
Under the rank-1 surrogates above with $\rho>0$, the branch residuals are:
\begin{align}
\boldsymbol{\tau}_{\mathrm{sgs},\mathbf{B}} &\;=\; \frac{1}{\rho}\!\left[\frac{\mathbf{G}\mathbf{A}^{\top}\mathbf{v}_{\!\mathbf{B}}\,\mathbf{v}_{\!\mathbf{B}}^{\top}}{\rho+\|\mathbf{v}_{\!\mathbf{B}}\|^2} - \frac{\mathbf{G}\mathbf{v}_f\,(\mathbf{A}\mathbf{v}_f)^{\top}}{\rho+\|\mathbf{v}_f\|^2}\right], \label{eq:method-tau-B}\\
\boldsymbol{\tau}_{\mathrm{sgs},\mathbf{A}} &\;=\; \frac{1}{\rho}\!\left[\frac{\mathbf{B}^{\top}\mathbf{G}\mathbf{v}_{\!\mathbf{A}}\,\mathbf{v}_{\!\mathbf{A}}^{\top}}{\rho+\|\mathbf{v}_{\!\mathbf{A}}\|^2} - \frac{\mathbf{B}^{\top}\mathbf{G}\mathbf{v}_f\,\mathbf{v}_f^{\top}}{\rho+\|\mathbf{v}_f\|^2}\right]. \label{eq:method-tau-A}
\end{align}
Proof in Appendix~\ref{sec:app-thm-52}.
\end{proposition}

Lemma~\ref{lem:chain-rule} and Proposition~\ref{prop:subgrid-residual} together identify $\boldsymbol{\tau}_{\mathrm{sgs}}$ as the second-order curvature gap that VMS-EO must inject. We now describe the corrected coarse update.

\paragraph{Rank-1 Surrogate for LoRA Branches.}
\label{sec:method-correction}

Let $\mathbf{s}_c:=\boldsymbol{\Sigma}_c\mathcal{R}(\mathbf{G})$ be the pure coarse preconditioned direction, which equals the LoRA-space LOREN direction without any VMS-EO correction. VMS-EO simply adds a branch-weighted subgrid correction $\mathbf{s}_c+\lambda\,\boldsymbol{\tau}_{\mathrm{sgs}}$, where $\lambda=0$ recovers LoRA-space LOREN and a larger $\lambda$ injects more fine-grid curvature feedback. The two branches carry independent weights $\lambda_{\!\mathbf{B}},\lambda_{\!\mathbf{A}}\in[0,1]$.

\paragraph{VMS-EO corrected updates.}
Let $\widehat{\mathbf{g}}_{c,\mathbf{B}}\in\mathbb R^{m\times r}$ and $\widehat{\mathbf{g}}_{c,\mathbf{A}}\in\mathbb R^{r\times n}$ denote the preconditioned RLOO directions for the two LoRA branches from Eq.~\eqref{eq:method-rloo}. With branch step sizes $\eta_{\!\mathbf B},\eta_{\!\mathbf A}>0$, the corrected updates are:
\begin{align}
\Delta\mathbf{B} &\;=\;-\eta_{\!\mathbf{B}}\bigl(\widehat{\mathbf{g}}_{c,\mathbf{B}}+\lambda_{\!\mathbf{B}}\,\boldsymbol{\tau}_{\mathrm{sgs},\mathbf{B}}\bigr), \label{eq:method-vms-B}\\
\Delta\mathbf{A} &\;=\;-\eta_{\!\mathbf{A}}\bigl(\widehat{\mathbf{g}}_{c,\mathbf{A}}+\lambda_{\!\mathbf{A}}\,\boldsymbol{\tau}_{\mathrm{sgs},\mathbf{A}}\bigr). \label{eq:method-vms-A}
\end{align}
Implementation details are given in Appendix~\ref{sec:app-surrogate}.

\paragraph{Adaptive mixing weights.}
Proposition~\ref{prop:theory-oracle} characterizes an oracle mixing weight that minimizes the estimation MSE. Since it depends on unknown correction and error statistics, we use normalized angular disagreement between coarse and fine curvature directions as a practical proxy for selecting $\lambda_{\!\mathbf{B}},\lambda_{\!\mathbf{A}}\in[0,1]$. For the B-branch we use the formula below, and the A-branch is analogous after replacing $\mathbf{v}_{\!\mathbf{B}}$ with $\mathbf{v}_{\!\mathbf{A}}$ and the projection $\mathbf{A}\mathbf{v}_f$ with $\mathbf{v}_f$:
\begin{equation}\label{eq:method-adaptive}
\delta_{\!\mathbf{B}}:=\frac{1}{2}\left\|\frac{\mathbf{v}_{\!\mathbf{B}}}{\|\mathbf{v}_{\!\mathbf{B}}\|}-\frac{\mathbf{A}\mathbf{v}_f}{\|\mathbf{A}\mathbf{v}_f\|}\right\|, \qquad
\lambda_{\!\mathbf{B}}=\mathrm{clip}(\delta_{\!\mathbf{B}},\lambda_{\min},\lambda_{\max}).
\end{equation}
When $\delta$ approaches zero the two curvature directions become aligned, so $\lambda$ falls to $\lambda_{\min}$ and the correction receives its minimum mixing weight. When $\delta$ approaches one the directions diverge, so $\lambda$ rises to $\lambda_{\max}$ and injects more fine-grid feedback.

\subsection{\Engine: V-cycle Coupling}
\label{sec:method-vcycle}

\Engine\ alternates between $M$-epoch coarse ($\boldsymbol{\phi}$) and $N$-epoch fine ($\mathbf{W}$) phases (Figure~\ref{fig:concept} Middle), updating the base $\mathbf{W}_0$ in $\mathbf{W}=\mathbf{W}_0+\mathbf{B}\mathbf{A}$ only at phase boundaries. A cyclic schedule couples them, using $\mathcal{P}$ and $\mathcal{R}$ operators to transfer parameters, momentum, and curvature.

\paragraph{Two-phase schedule.}
\label{sec:method-coarse-phase}
\emph{Coarse phase.} \Engine\ updates $\boldsymbol{\phi}=(\mathbf{B},\mathbf{A})$ via Eqs.~\eqref{eq:method-vms-B} and \eqref{eq:method-vms-A} and uses~\eqref{eq:method-rloo} to take one curvature step per branch with learning rate $\nu>0$ (Algorithm~\ref{alg:engine-main}, line 6). Crucially, the $K$ RLOO samples $\{u_k,f_k\}$ are reused from those already drawn for $(\Delta\mathbf{B},\Delta\mathbf{A})$, so this curvature step costs no additional forward passes. The operator $\widetilde{\mathbf{H}}=\widetilde{\mathbf{H}}(\boldsymbol{\theta},\mathbf{v}_f)$ is endogenous, however, so $\mathbf{v}_f$ goes stale across coarse steps and causes $\boldsymbol{\tau}_{\mathrm{sgs}}$ to inject outdated information through its $\mathbf{v}_f$ dependence. We therefore trigger a Hessian refresh every $S$ coarse steps as illustrated in Algorithm~\ref{alg:engine-main}, where $K$ extra full-space perturbations update $\mathbf{v}_f$ using~\eqref{eq:method-rloo} with learning rate $\nu$ together with the projection scalars in Proposition~\ref{prop:subgrid-residual} while $\boldsymbol{\theta}$, $\mathbf{B}$, and $\mathbf{A}$ remain unchanged. This refresh-and-reuse strategy parallels Lazy Hessian~\citep{doikov2023second}, which also recomputes curvature every $S$ steps and reuses it in between to cut arithmetic cost without sacrificing convergence.

\emph{Fine phase.} \Engine\ returns to the full parameter space and applies $\boldsymbol{\Sigma}_f$-preconditioned LOREN steps from Eq.~\eqref{eq:method-rloo} to $\mathbf{W}$, simultaneously refreshing $\mathbf{v}_f$ for the next coarse phase.

\paragraph{Prolongation $\mathcal{P}$ from coarse to fine.}
\label{sec:method-prolongation}
The coarse update $\mathbf{B}\mathbf{A}$ is absorbed into the base weight $\mathbf{W}_0$. We save the current $\mathbf{A}$ as $\mathbf{A}_{\mathrm{old}}$ before reinitialization. This cached matrix maps vectors from $\mathbb{R}^n$ to $\mathbb{R}^r$ during the next transition from fine to coarse. Branch momenta combine into $\mathbf{m}_f$ via the LoRA product rule, and curvature is EMA-blended with weight $\beta$. Algorithm~\ref{alg:engine-main} specifies the transfers.

\paragraph{Restriction $\mathcal{R}$ from fine to coarse.}
\label{sec:method-restriction}
We set the base weight $\mathbf{W}_0$ to $\mathbf{W}_{\mathrm{fine\_best}}$, rescale $\mathbf{B}$, and reset $\mathbf{A}$ to zero so that $\mathbf{W}_0+\mathbf{B}\mathbf{A}=\mathbf{W}_0$ holds exactly at cycle start. Momentum is projected onto the two branches via $\mathcal{R}$ with default damping factor $\alpha=0.5$, and the two branch curvatures use the cached basis $\mathbf{A}_{\mathrm{old}}$. Algorithm~\ref{alg:engine-main} gives these transfer formulas.

\Engine\ maintains only the LoRA factors $(\mathbf{B},\mathbf{A})$ and the rank-1 curvature vectors $(\mathbf{v}_f,\mathbf{v}_{\!\mathbf{B}},\mathbf{v}_{\!\mathbf{A}})$, never explicitly storing a full Hessian. Algorithm~\ref{alg:engine-main} summarizes the procedure, and implementation details are in Appendix~\ref{sec:app-vms-eo-details}.

\begin{algorithm}[t]
\caption{\Engine\ (see Appendix~\ref{sec:app-vms-eo-details} for details).}
\label{alg:engine-main}
\begin{algorithmic}[1]
\Require frozen $\mathbf{W}$, dataset $\mathcal{D}$, rank $r$, cycles $V$, coarse/fine epochs $M,N$, queries $K$, smoothing $\epsilon$, damping $\rho$, learning rates $\eta_{\!\mathbf{B}},\eta_{\!\mathbf{A}},\eta_f,\nu$, mixing range $[\lambda_{\min},\lambda_{\max}]$, transfer weight $\beta$, fine-probe interval $S$ (every $S$ coarse steps, refresh $\mathbf{v}_f$ and $\lambda$).
\State Initialize $\mathbf{B}$ randomly, set $\mathbf{A}\!\leftarrow\!\mathbf{0}$, and initialize $\mathbf{v}_{\!\mathbf{B}},\mathbf{v}_{\!\mathbf{A}},\mathbf{v}_f$.
\For{$v=1,\ldots,V$}
    \State \textcolor{engineorange}{\textit{Coarse phase ($M$ epochs in $\boldsymbol{\phi}=(\mathbf{B},\mathbf{A})$):}} \Comment{\textcolor{engineorange}{$\blacksquare$}~Coarse space}
    \For{$t=1,\ldots,M\!\cdot\!\mathrm{steps/epoch}$}
        \State \textcolor{engineorange}{Draw $K$ LoRA perturbations and estimate $\widehat{\mathbf{g}}_{c,\mathbf{B}},\widehat{\mathbf{g}}_{c,\mathbf{A}}$ via~\eqref{eq:method-rloo}.}
        \State \textcolor{engineorange}{$\mathbf{v}_{\!\mathbf{B}}\!\leftarrow\!\mathbf{v}_{\!\mathbf{B}}\!-\!\nu g(\mathbf{v}_{\!\mathbf{B}})$, $\mathbf{v}_{\!\mathbf{A}}\!\leftarrow\!\mathbf{v}_{\!\mathbf{A}}\!-\!\nu g(\mathbf{v}_{\!\mathbf{A}})$ on the same $K$ samples.}
        \If{$t\bmod S=0$}
            \State \textcolor{enginegreen}{$\mathbf{v}_f\!\leftarrow\!\mathbf{v}_f\!-\!\nu g(\mathbf{v}_f)$ via~\eqref{eq:method-rloo}.} \Comment{\textcolor{enginegreen}{$\blacksquare$}~Hessian refresh, $K$ extra fwd}
            \State \textcolor{enginegreen}{Update $\lambda_{\!\mathbf{B}},\lambda_{\!\mathbf{A}}$ via \eqref{eq:method-adaptive}.}
        \EndIf
        \State \textcolor{enginepurple}{Compute $\boldsymbol{\tau}_{\mathrm{sgs}}$ via Prop.~\ref{prop:subgrid-residual} (Eqs.~\eqref{eq:method-tau-B} and \eqref{eq:method-tau-A}).} \Comment{\textcolor{enginepurple}{$\blacksquare$}~VMS-EO}
        \State \textcolor{enginepurple}{Update $(\mathbf{B},\mathbf{A})$ via Eqs.~\eqref{eq:method-vms-B} and \eqref{eq:method-vms-A}.}
    \EndFor
    \State \textit{Prolongation $\mathcal{P}$ ($\mathbf{C}\!\to\!\mathbf{F}$):}\ $\mathbf{W}_0\!\leftarrow\!\mathbf{W}_0\!+\!\mathbf{B}\mathbf{A}$,\ $\mathbf{A}_{\mathrm{old}}\!\leftarrow\!\mathbf{A}$,\ $\mathbf{m}_f\!=\!\mathbf{m}_{\!\mathbf{B}}\mathbf{A}\!+\!\mathbf{B}\mathbf{m}_{\!\mathbf{A}}$,\ $\mathbf{v}_f\!\leftarrow\!\mathrm{normalize}((1\!-\!\beta)\mathbf{v}_f\!+\!\beta\,\mathrm{normalize}(\mathbf{A}^{\!\top}\mathbf{v}_{\!\mathbf{B}}\!+\!\mathbf{v}_{\!\mathbf{A}}))$.
    \State \textcolor{engineblue}{\textit{Fine phase ($N$ epochs in $\mathbf{W}$):}\ LOREN on $\mathbf{W}$ with lr $\eta_f$, $K$ queries via~\eqref{eq:method-rloo}, updating $\mathbf{v}_f$.} \Comment{\textcolor{engineblue}{$\blacksquare$}~Fine space}
    \State \textit{Restriction $\mathcal{R}$ ($\mathbf{F}\!\to\!\mathbf{C}$):}\ $\mathbf{W}_0\!\leftarrow\!\mathbf{W}_{\mathrm{fine\_best}}$,\ rescale $\mathbf{B}$, $\mathbf{A}\!\leftarrow\!\mathbf{0}$,\ $\mathbf{m}_{\!\mathbf{A}}\!=\!\alpha\mathbf{B}^{\!\top}\mathbf{m}_f$, $\mathbf{m}_{\!\mathbf{B}}\!=\!\alpha\mathbf{m}_f\mathbf{A}_{\mathrm{old}}^{\!\top}$,\ $\mathbf{v}_{\!\mathbf{A}}\!=\!\mathrm{normalize}(\mathbf{v}_f)$, $\mathbf{v}_{\!\mathbf{B}}\!=\!\mathrm{normalize}(\mathbf{A}_{\mathrm{old}}\mathbf{v}_f)$.
\EndFor
\State \Return $\mathbf{W}$.
\end{algorithmic}
\end{algorithm}

\subsection{Statistical and Convergence Analysis}
\label{sec:method-statistical}

We provide a detailed analysis of the trade-off between bias and variance in \Engine. Under appropriate assumptions, coarse sampling reduces the dimension-dependent variance bound and exact correction eliminates structural bias in the local-linear limit. We also establish convergence bounds for coarse updates.

Fix the current weights, LoRA factors, and covariances. Subscripts $f$ and $c$ denote fine and coarse spaces. In row-wise vector coordinates, let $\mathbf{g}_f^{\mathrm{raw}}$ be the full-space gradient and $\mathbf{g}_c^{\mathrm{raw}}=\mathcal{R}(\mathbf{g}_f^{\mathrm{raw}})$ the gradient with respect to $\boldsymbol\phi$. The coarse direction is $\mathbf{s}_c=\boldsymbol\Sigma_c\mathbf{g}_c^{\mathrm{raw}}$, and the common comparison target is $\mathbf{g}_\star=\mathcal{R}(\boldsymbol\Sigma_f\mathbf{g}_f^{\mathrm{raw}})$. Thus Eq.~\eqref{eq:method-tau-sgs} gives $\boldsymbol\tau_{\mathrm{sgs}}=\mathbf{g}_\star-\mathbf{s}_c$.

We use a Gaussian estimator with $b$ independent directions and smoothing radius $\epsilon$. Write $F(\boldsymbol\phi)=f(\mathbf{W}_0+\mathbf{B}\mathbf{A})$ and draw $\mathbf{z}_i\sim\mathcal{N}(\mathbf{0},\boldsymbol\Sigma_c)$ independently. Define:
\begin{equation}\label{eq:method-gaussian-direction}
\widehat{\mathbf{g}}_{c,\epsilon,b}
=\frac{1}{b}\sum_{i=1}^{b}
\frac{F(\boldsymbol\phi+\epsilon\mathbf{z}_i)-F(\boldsymbol\phi)}{\epsilon}\,\mathbf{z}_i.
\end{equation}
Here $\mathbf z_i=[\mathbf z_{\mathbf B,i}^{\top},\mathbf z_{\mathbf A,i}^{\top}]^{\top}\in\mathbb R^{d_c}$ stacks B- and A-branch perturbations of dimensions $mr$ and $rn$, with $d_c=r(m+n)$. The covariance $\boldsymbol\Sigma_c=\operatorname{diag}(\boldsymbol\Sigma_{\!\mathbf B},\boldsymbol\Sigma_{\!\mathbf A})\in\mathbb R^{d_c\times d_c}$ has blocks of sizes $mr\times mr$ and $rn\times rn$, respectively. Implementation uses RLOO from Eq.~\eqref{eq:method-rloo}. The next result isolates the effect of exact correction in the local-linear limit, where the smoothing bias vanishes.

\begin{proposition}[Common-target error with exact correction]
\label{prop:method-common-target}
Suppose $F$ has a Lipschitz continuous gradient and $\boldsymbol\Sigma_c$ is fixed and positive definite. As $\epsilon\to0$ with $b$ fixed, Eq.~\eqref{eq:method-gaussian-direction} converges in mean square to $\widehat{\mathbf{g}}_{c,0,b}=b^{-1}\sum_i\mathbf{z}_i\mathbf{z}_i^\top\mathbf{g}_c^{\mathrm{raw}}$. Its mean is $\mathbf{s}_c$ and its trace variance is:
\begin{equation}\label{eq:method-local-variance}
V_{c,0}
=\frac{\operatorname{tr}(\boldsymbol\Sigma_c)(\mathbf{g}_c^{\mathrm{raw}})^\top\boldsymbol\Sigma_c\mathbf{g}_c^{\mathrm{raw}}
+(\mathbf{g}_c^{\mathrm{raw}})^\top\boldsymbol\Sigma_c^2\mathbf{g}_c^{\mathrm{raw}}}{b}.
\end{equation}
With exact deterministic correction and both branch weights equal to one, the common-target errors satisfy:
\begin{align}
\mathbb{E}\|\widehat{\mathbf{g}}_{c,0,b}-\mathbf{g}_\star\|^2
&=V_{c,0}+\|\boldsymbol\tau_{\mathrm{sgs}}\|^2,\label{eq:method-coarse-mse}\\
\mathbb{E}\|\widehat{\mathbf{g}}_{c,0,b}+\boldsymbol\tau_{\mathrm{sgs}}-\mathbf{g}_\star\|^2
&=V_{c,0}.\label{eq:method-ideal-mse}
\end{align}
Theorem~\ref{prop:theory-rge} proves the stated mean-square limit of the Gaussian estimator $\widehat{\mathbf g}_{c,\epsilon,b}$. Lemma~\ref{lem:theory-linear-covariance} gives Eq.~\eqref{eq:method-local-variance}, and Proposition~\ref{prop:theory-mse} yields Eqs.~\eqref{eq:method-coarse-mse} and \eqref{eq:method-ideal-mse} in the limit. Proof in Appendix~\ref{sec:theory-local-comparison}.
\end{proposition}

\textbf{Variance reduction and bias correction.}
For isotropic sampling with $\boldsymbol\Sigma_c=I_{d_c}$, Eq.~\eqref{eq:method-local-variance} reduces to $V_{c,0}=\frac{d_c+1}{b}\|\mathbf g_c^{\mathrm{raw}}\|^2$, compared with the fine-space factor $\frac{d_f+1}{b}$. For anisotropic sampling with fixed positive-definite $\boldsymbol\Sigma_c$ and $\|\boldsymbol\Sigma_c\|_{\mathrm{op}}\le M_c$, the same equation gives the upper bound $V_{c,0}\le\frac{d_c+1}{b}M_c^2\|\mathbf g_c^{\mathrm{raw}}\|^2$. Under comparable gradient and spectral scales, $d_c\ll d_f$ reduces the dimension-dependent variance bound. In the local-linear limit, Proposition~\ref{prop:theory-mse} shows that exact correction eliminates the structural bias relative to $\mathbf g_\star$ while preserving the coarse sampling variance, decreasing MSE by $\|\boldsymbol\tau_{\mathrm{sgs}}\|^2$. Ideal \Engine\ thus combines coarse-space variance reduction with structural-bias elimination. Appendix~\ref{sec:theory-mse} gives finite-radius bounds and a common-target comparison in Table~\ref{tab:theory-common-target}.

For finite smoothing radius $\epsilon$ and $L_c$-smooth $F$, Theorem~\ref{prop:theory-rge} in Appendix~\ref{sec:theory-gaussian} gives $\|\mathbf b_{c,\epsilon}\|=\mathcal O(\epsilon\sqrt{d_c})$ with $L_c$ and $M_c$ held fixed. Here $\mathbf b_{c,\epsilon}=\mathbb E\widehat{\mathbf g}_{c,\epsilon,b}-\mathbf s_c$ is the residual smoothing bias. Proposition~\ref{prop:theory-mse} gives the MSE reduction $\|\boldsymbol\tau_{\mathrm{sgs}}\|^2-2\langle\mathbf b_{c,\epsilon},\boldsymbol\tau_{\mathrm{sgs}}\rangle$, quantifying the effect of this remaining bias. For approximate corrections, Proposition~\ref{prop:theory-noisy-mse} accounts for their bias, variance, and dependence on the coarse estimate. Proposition~\ref{prop:theory-projection} in Appendix~\ref{sec:theory-projection} separates row compression error from scalar coefficient mismatch.

\textbf{Convergence analysis.}
Consider the ideal Gaussian update $\boldsymbol\phi_{t+1}=\boldsymbol\phi_t-\eta(\widehat{\mathbf g}_{c,\epsilon,b,t}+\boldsymbol\tau_{\mathrm{sgs},t})$ on a fixed coarse objective $F$. Under the assumptions in Appendix~\ref{sec:theory-convergence}, Theorem~\ref{prop:theory-convergence} gives the following bound for the prescribed $\eta=\Theta(T^{-1/2})$:
\begin{equation}\label{eq:method-convergence-rate}
\frac{1}{T}\sum_{t=0}^{T-1}
\mathbb E\|\nabla F(\boldsymbol\phi_t)\|^2
\le O(T^{-1/2})+O(\epsilon^2).
\end{equation}
Here $T$ counts iterations. The bound separates optimization decay from the smoothing residual, with coarse-dimensional sampling constants. Corollary~\ref{cor:theory-approximate-rate} extends the bound to approximate corrections, with correction bias affecting the residual and correction variance affecting the sampling constants.

\section{Experiments}
\label{sec:experiments}
We evaluate \Engine\ on 16 (model, task) combinations, assessing accuracy, efficiency, and the contributions of V-cycle and VMS-EO.

\subsection{Experimental Setup and Main Results}
\label{sec:exp-setup}

\paragraph{Setup.}
We evaluate four autoregressive LLMs of size 1.5B to 13B, namely GPT-2-XL~\citep{radford2019language}, OPT-2.7B and OPT-13B~\citep{zhang2022optopenpretrainedtransformer}, and LLaMA-3-8B~\citep{grattafiori2024llama}, on 16 prompt-free classification tasks drawn from GLUE~\citep{wang2018glue} and SuperGLUE~\citep{wang2019superglue}. We compare \Engine\ against LOREN together with five other ZO baselines, whose full list and background appear in Appendix~\ref{sec:related-work}. All baselines are independently reproduced under LOREN's setup. On top of this protocol, \Engine\ introduces only two new hyperparameters. The C60F60 V-cycle alternates 60 coarse epochs with 60 fine epochs, and the curvature refresh interval $S$ is set per (model, task) combination as listed in Table~\ref{tab:engine-hyperparameters}. The LoRA rank $r$ scales with model size. We report mean accuracy (\%) over five seeds for all tasks. See Appendices~\ref{sec:app-tables} and~\ref{sec:app-exp-setup} for details.

\begin{figure}[t]
\centering
\includegraphics[width=0.92\linewidth]{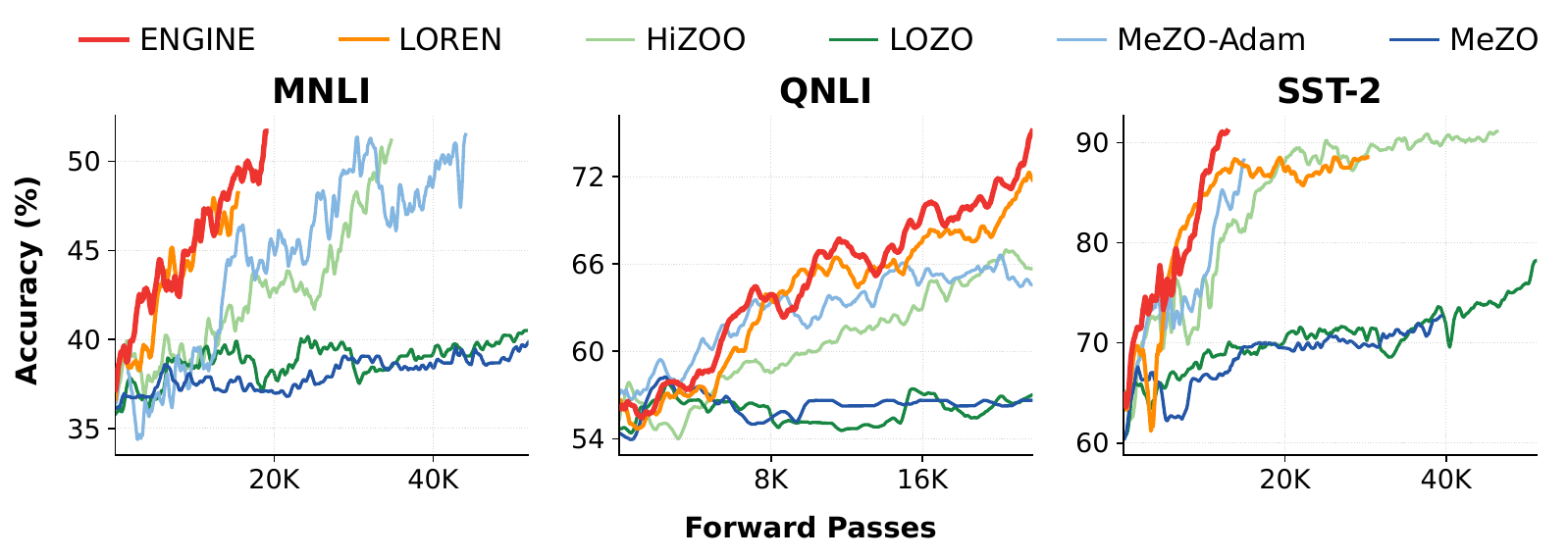}
\caption{\textbf{Visualization of validation accuracy on GPT-2-XL across three GLUE tasks (MNLI, QNLI, SST-2).} \Engine\ reaches higher peak accuracy than all baselines on every task. See Appendix~\ref{sec:app-learning-curves} for training loss curves and all four models.}
\label{fig:exp-convergence}
\end{figure}

\paragraph{Main results.}
Table~\ref{tab:main-results} and Figure~\ref{fig:bars-main} show that \Engine\ outperforms LOREN and other baseline methods on every (model, task) cell. Representative training curves appear in Figure~\ref{fig:exp-convergence}.

\begin{table}[t]
\centering
\caption{Main results using accuracy (\%) for all tasks. \textbf{Bold} and \textcolor{secondcyan}{\textbf{cyan}} mark the best and second best per column. Highlighted rows give \Engine\ and its absolute gain over LOREN. Full results in Appendix~\ref{sec:app-main-tables}.}
\label{tab:main-results}
\setlength{\tabcolsep}{4pt}
\renewcommand{\arraystretch}{1.05}
\small
\begin{tabular}{l ccccc ccccc}
\toprule
\textbf{GLUE} (FP32) & \multicolumn{5}{c}{\textbf{GPT-2-XL} (1.5B)} & \multicolumn{5}{c}{\textbf{OPT-2.7B}} \\
\cmidrule(lr){2-6}\cmidrule(lr){7-11}
Method & MNLI & QNLI & SST-2 & CoLA & Avg. & MNLI & QNLI & SST-2 & CoLA & Avg. \\
\midrule
MeZO        & 38.4 & 59.2 & 74.3 & 65.8 & 59.4 & 50.4 & \snd{75.2} & \snd{88.7} & 68.4 & 70.7 \\
MeZO-Adam   & \snd{50.8} & 71.9 & \snd{91.2} & 71.4 & 71.3 & 45.9 & 72.3 & 86.9 & \snd{68.5} & 68.4 \\
MeZO-SVRG   & 49.8 & 65.5 & 84.5 & 65.6 & 66.4 & 41.5 & 59.8 & 62.9 & 66.5 & 57.7 \\
LOZO        & 42.9 & 60.6 & 80.0 & 66.2 & 62.4 & 43.2 & 59.5 & 66.2 & 66.3 & 58.8 \\
HiZOO       & 48.4 & 66.8 & 88.9 & 71.2 & 68.8 & 50.9 & 74.7 & 88.2 & 66.3 & 70.0 \\
LOREN       & 50.6 & \snd{74.1} & 89.0 & \snd{72.0} & \snd{71.4} & \snd{54.1} & 72.6 & 88.2 & 68.1 & \snd{70.8} \\
\midrule
\rowcolor{enginerow}
\textbf{\Engine} (Ours) & \best{52.0} & \best{77.0} & \best{91.8} & \best{72.7} & \best{73.4} & \best{55.5} & \best{75.4} & \best{91.0} & \best{69.5} & \best{72.9} \\
\rowcolor{enginerow}
$\Delta$ vs LOREN & \gain{1.4} & \gain{2.9} & \gain{2.8} & \gain{0.7} & \gain{2.0} & \gain{1.4} & \gain{2.8} & \gain{2.8} & \gain{1.4} & \gain{2.1} \\
\midrule
\textbf{SuperGLUE} (BF16) & \multicolumn{5}{c}{\textbf{LLaMA-3-8B}} & \multicolumn{5}{c}{\textbf{OPT-13B}} \\
\cmidrule(lr){2-6}\cmidrule(lr){7-11}
Method & RTE & BoolQ & WiC & CB & Avg. & RTE & BoolQ & WiC & CB & Avg. \\
\midrule
MeZO        & \snd{59.1} & 64.0 & \snd{59.3} & 71.5 & 63.5 & 59.0 & 64.4 & 57.1 & 72.7 & 63.3 \\
MeZO-Adam   & 58.7 & 63.2 & 57.1 & 68.9 & 62.0 & \snd{60.8} & 63.4 & 55.7 & 70.2 & 62.2 \\
MeZO-SVRG   & 58.2 & 63.1 & 58.0 & 64.9 & 61.1 & 59.3 & \best{65.2} & 56.8 & 69.4 & 62.7 \\
LOZO        & 57.8 & 63.6 & 57.2 & 71.1 & 62.4 & 58.2 & 63.4 & \snd{57.7} & 68.8 & 62.0 \\
HiZOO       & 58.2 & 63.6 & 59.2 & 67.1 & 62.0 & 59.5 & 63.7 & 57.0 & 72.7 & 63.2 \\
LOREN       & 58.4 & \snd{64.5} & 58.9 & \snd{72.4} & \snd{63.6} & 59.5 & 63.2 & 57.0 & \snd{73.7} & \snd{63.4} \\
\midrule
\rowcolor{enginerow}
\textbf{\Engine} (Ours) & \best{59.8} & \best{66.0} & \best{59.8} & \best{73.2} & \best{64.7} & \best{62.1} & \snd{65.2} & \best{59.0} & \best{79.1} & \best{66.4} \\
\rowcolor{enginerow}
$\Delta$ vs LOREN & \gain{1.4} & \gain{1.5} & \gain{0.9} & \gain{0.8} & \gain{1.1} & \gain{2.6} & \gain{2.0} & \gain{2.0} & \gain{5.4} & \gain{3.0} \\
\bottomrule
\end{tabular}
\end{table}

\paragraph{Efficiency and memory.}
Figure~\ref{fig:speedup-convergence} further shows that on every setting, \Engine\ reaches LOREN's best accuracy with substantially fewer forward passes and less wall-clock time, achieving up to a $2.61\times$ forward-pass speed-up and a $2.58\times$ wall-clock speed-up on LLaMA-3-8B (CB). Meanwhile, \Engine's peak GPU memory matches LOREN's, because the fine phase is equivalent to vanilla LOREN and determines the peak, while the coarse phase only stores LoRA-scale optimizer and curvature states. Table~\ref{tab:common-target-efficiency} compares all methods at common target accuracies.

\subsection{Ablation: Where does \Engine's gain originate from?}
\label{sec:exp-ablation}

\paragraph{Setup and design.}
We compare LOREN, \Engine\ w/o VMS-EO, and \Engine\ on the same 16 settings, with other settings matched (Appendix~\ref{sec:app-exp-setup}). The variant without VMS-EO retains the V-cycle and disables the correction. Its comparison with LOREN isolates the V-cycle's contribution, while its comparison with full \Engine\ isolates the correction's contribution to performance.

\begin{table}[t]
\centering
\caption{Ablation accuracy (\%) on all 16 settings. \textbf{Bold} marks the best per column, and \textcolor{secondcyan}{\textbf{cyan}} marks the second best. The \colorbox{enginerow}{\strut highlighted row} is our method, with the green \textcolor{gaingreen}{$\Delta$ vs LOREN} row showing the absolute gain. Full results in Appendix~\ref{sec:app-tables}.}
\label{tab:ablation-vmseo}
\setlength{\tabcolsep}{4pt}
\renewcommand{\arraystretch}{1.05}
\small
\begin{tabular}{l ccccc ccccc}
\toprule
\textbf{GLUE} (FP32) & \multicolumn{5}{c}{\textbf{GPT-2-XL} (1.5B)} & \multicolumn{5}{c}{\textbf{OPT-2.7B}} \\
\cmidrule(lr){2-6}\cmidrule(lr){7-11}
Method & MNLI & QNLI & SST-2 & CoLA & Avg. & MNLI & QNLI & SST-2 & CoLA & Avg. \\
\midrule
LOREN                 & \snd{50.6} & \snd{74.1} & 89.0 & \snd{72.0} & \snd{71.4} & 54.1 & 72.6 & 88.2 & 68.1 & 70.8 \\
\Engine\ w/o VMS-EO     & 48.8 & 72.7 & \snd{90.2} & 71.5 & 70.8 & \snd{54.7} & \snd{73.8} & \snd{89.5} & \snd{69.1} & \snd{71.8} \\
\midrule
\rowcolor{enginerow}
\textbf{\Engine} (Ours) & \best{52.0} & \best{77.0} & \best{91.8} & \best{72.7} & \best{73.4} & \best{55.5} & \best{75.4} & \best{91.0} & \best{69.5} & \best{72.9} \\
\rowcolor{enginerow}
$\Delta$ vs LOREN & \gain{1.4} & \gain{2.9} & \gain{2.8} & \gain{0.7} & \gain{2.0} & \gain{1.4} & \gain{2.8} & \gain{2.8} & \gain{1.4} & \gain{2.1} \\
\midrule
\textbf{SuperGLUE} (BF16) & \multicolumn{5}{c}{\textbf{LLaMA-3-8B}} & \multicolumn{5}{c}{\textbf{OPT-13B}} \\
\cmidrule(lr){2-6}\cmidrule(lr){7-11}
Method & RTE & BoolQ & WiC & CB & Avg. & RTE & BoolQ & WiC & CB & Avg. \\
\midrule
LOREN                 & \snd{58.4} & 64.5 & \snd{58.9} & \snd{72.4} & \snd{63.6} & \snd{59.5} & 63.2 & 57.0 & 73.7 & 63.4 \\
\Engine\ w/o VMS-EO     & 58.2 & \snd{65.2} & 57.8 & 71.4 & 63.2 & 59.0 & \snd{64.1} & \snd{57.8} & \snd{75.0} & \snd{64.0} \\
\midrule
\rowcolor{enginerow}
\textbf{\Engine} (Ours) & \best{59.8} & \best{66.0} & \best{59.8} & \best{73.2} & \best{64.7} & \best{62.1} & \best{65.2} & \best{59.0} & \best{79.1} & \best{66.4} \\
\rowcolor{enginerow}
$\Delta$ vs LOREN & \gain{1.4} & \gain{1.5} & \gain{0.9} & \gain{0.8} & \gain{1.1} & \gain{2.6} & \gain{2.0} & \gain{2.0} & \gain{5.4} & \gain{3.0} \\
\bottomrule
\end{tabular}
\end{table}

\subsection{Analysis: Attributing the Dual-Scale Benefits of \Engine}
\label{sec:exp-analysis}

\paragraph{Result.}
Table~\ref{tab:ablation-vmseo} yields two observations. First, \Engine\ w/o VMS-EO has a mix of wins and losses against LOREN across the 16 settings. Second, \Engine\ achieves higher accuracy than both LOREN and \Engine\ w/o VMS-EO across all four models. Table~\ref{tab:app-ablation-full} gives the full mean$\pm$std results, and Figure~\ref{fig:app-ablation-4x4} shows the ablation convergence curves.

\paragraph{Analysis.}
The two observations identify distinct contributions.
\begin{itemize}[leftmargin=1.5em,itemsep=2pt,topsep=2pt]
  \item \textit{The V-cycle alone achieves mixed gains against LOREN.} Coarse sampling in $d_c\ll d_f$ dimensions reduces the variance bound under comparable gradient and spectral scales (Proposition~\ref{prop:method-common-target}). The second-order restriction mismatch in Proposition~\ref{prop:subgrid-residual} biases the uncorrected coarse direction and can offset this variance benefit. Alternating fine phases retain full-space curvature-aware optimization, while coarse phases reduce per-step cost.
  \item \textit{Incorporating VMS-EO yields consistent performance gains.} By feeding full-space curvature information back into the subspace through Eqs.~\eqref{eq:method-vms-B} and \eqref{eq:method-vms-A}, VMS-EO corrects the structural bias. Proposition~\ref{prop:method-common-target} shows that exact correction eliminates this bias without increasing coarse sampling variance in the local-linear limit. Combining coarse sampling with curvature correction thus helps explain \Engine's gains over LOREN and the uncorrected V-cycle.
\end{itemize}

\section{Conclusions}
\label{sec:conclusions}
In this paper, we propose \Engine, a zeroth-order LLM fine-tuning framework inspired by Variational MultiScale (VMS) theory from computational mechanics. From the VMS perspective, we treat the LoRA parameter space as the coarse space and the full parameter space as the fine space. By reformulating VMS with an endogenous, state-dependent Hessian, we derive the VMS-EO subgrid correction to incorporate fine-space curvature information into coarse updates. \Engine\ integrates this correction into an alternating V-cycle that couples the two spaces through mutual feedback. Our statistical analysis shows that, under the stated assumptions, coarse sampling reduces the dimension-dependent variance bound and exact subgrid correction eliminates curvature-mismatch bias in the local-linear limit. We further derive nonconvex stationarity bounds for corrected coarse updates. Across 16 model-task settings, \Engine\ consistently outperforms state-of-the-art ZO baselines, achieving substantial forward-pass and wall-clock speed-ups with no additional peak memory overhead. More broadly, this work provides a multiscale perspective on optimization, showing how efficient low-dimensional updates can be guided by full-space curvature information. We hope this perspective inspires further connections between multiscale modeling and optimization. Appendix~\ref{sec:app-limitations} discusses limitations and future work.

\clearpage
\begin{ack}
We gratefully acknowledge the support of Westlake University Research Center for Industries of the Future
and Westlake University Center for High-performance Computing.

The authors declare no competing interests.
\end{ack}

{
\small
\bibliographystyle{plainnat}
\bibliography{references}
}

\newpage
\appendix

\startcontents[appendices]
\begingroup
\setlength{\parskip}{0pt}%
\printcontents[appendices]{l}{1}{%
  \section*{Appendix --- Table of Contents}%
  \setcounter{tocdepth}{2}%
}
\endgroup
\newpage

\section{Related Work}
\label{sec:related-work}

\Engine\ connects research on zeroth-order optimization, low-rank adaptation, and Variational MultiScale theory. These three lines of research motivate its sampling strategy, parameter spaces, and correction mechanism.

\paragraph{Zeroth-order LLM fine-tuning.}
Zeroth-order methods estimate gradients from function evaluations and reduce the memory required for LLM fine-tuning. MeZO uses in-place parameter perturbations to achieve memory usage comparable to inference. Its gradient estimates have high variance, which can slow optimization. MeZO-SVRG~\citep{gautam2024variance} uses periodic reference gradients to reduce this variance, with additional memory for stored parameters and gradients. LOZO~\citep{chen2024enhancing} exploits the low-rank structure of gradients to reduce the dimension of gradient estimation. Curvature information provides another way to improve the search direction. HiZOO~\citep{zhao2025second} estimates the diagonal Hessian and uses it to precondition the update. LOREN uses a low-rank Kronecker approximation to incorporate curvature into perturbation sampling. \Engine\ builds on this curvature representation, combining LoRA sampling with full-parameter updates. Its subgrid correction injects full-space curvature information into coarse updates.

\paragraph{Low-rank adaptation.}
LoRA represents a weight update as the product of two small matrices and optimizes their entries while keeping the base weight fixed. This reduces the number of trainable parameters and provides the coarse coordinates used by \Engine. ReLoRA~\citep{lialin2024relora} periodically merges the low-rank update into the base weight and reinitializes the low-rank factors. Successive updates can therefore accumulate in the full weight matrix. ReLoRA uses this procedure within first-order training. \Engine\ combines parameter merging with alternating LoRA and full-space ZO updates, transferring momentum and curvature between the two spaces.

\paragraph{Variational MultiScale theory.}
Variational MultiScale (VMS) theory originates in computational mechanics. It separates a problem into resolved coarse scales and unresolved fine scales. The effect of the fine scales enters the coarse equation through a subgrid correction. In the classical linear formulation, the governing operator is fixed by the physical problem and determines this correction. \Engine\ brings this perspective into zeroth-order optimization. The LoRA coordinates define the coarse space, and the correction accounts for curvature information lost when forming the coarse update. The curvature operator is estimated online and evolves during optimization. This state dependence motivates VMS with an Endogenous Operator.

\section{Structural Derivations and Proofs}
\label{sec:app-proofs}

In this appendix, we prove the LoRA chain rule and derive the subgrid residual formulas.

\subsection{Proof of Lemma~\ref{lem:chain-rule} (LoRA chain rule)}
\label{sec:app-lemma-51}

\begin{proof}
With $\mathbf W_0$ fixed, differentiating $\mathbf W=\mathbf W_0+\mathbf B\mathbf A$ gives:
\[
\mathrm d\mathbf W=\mathrm d\mathbf B\,\mathbf A+\mathbf B\,\mathrm d\mathbf A.
\]
The Frobenius inner product and cyclic trace identity give:
\[
\begin{aligned}
\mathrm df
&=\langle\nabla_{\!\mathbf W}f,\mathrm d\mathbf W\rangle_F\\
&=\langle(\nabla_{\!\mathbf W}f)\mathbf A^\top,\mathrm d\mathbf B\rangle_F
+\langle\mathbf B^\top\nabla_{\!\mathbf W}f,\mathrm d\mathbf A\rangle_F.
\end{aligned}
\]
Matching the coefficients of the independent differentials gives the components of $\mathcal R(\nabla_{\!\mathbf W}f)$:
\begin{align*}
\frac{\partial f}{\partial\mathbf B}&=(\nabla_{\!\mathbf W}f)\mathbf A^\top,\\
\frac{\partial f}{\partial\mathbf A}&=\mathbf B^\top\nabla_{\!\mathbf W}f.\qedhere
\end{align*}
\end{proof}

\subsection{Proof of Proposition~\ref{prop:subgrid-residual} (subgrid residual)}
\label{sec:app-thm-52}

\begin{proof}
For $\rho>0$ and $\mathbf v\in\mathbb R^d$, the Sherman-Morrison identity gives:
\begin{equation}
(\rho\mathbb I_d+\mathbf v\mathbf v^\top)^{-1}
=\frac1\rho\left(\mathbb I_d-\frac{\mathbf v\mathbf v^\top}{\rho+\|\mathbf v\|^2}\right).
\tag{B.1}\label{eq:app-sm-inverse}
\end{equation}
The Kronecker structure of $\boldsymbol\Sigma_f$ applies $(\rho\mathbb I_n+\mathbf v_f\mathbf v_f^\top)^{-1}$ independently to each row of $\mathbf G$. Equation~\eqref{eq:app-sm-inverse} therefore gives:
\[
\boldsymbol\Sigma_f\mathbf G
=\frac1\rho\left(\mathbf G-\frac{\mathbf G\mathbf v_f\mathbf v_f^\top}{\rho+\|\mathbf v_f\|^2}\right).
\]
Applying $\mathcal R_{\!\mathbf B}(\mathbf X)=\mathbf X\mathbf A^\top$ yields:
\[
\mathcal R_{\!\mathbf B}(\boldsymbol\Sigma_f\mathbf G)
=\frac1\rho\left(\mathbf G\mathbf A^\top-\frac{\mathbf G\mathbf v_f(\mathbf A\mathbf v_f)^\top}{\rho+\|\mathbf v_f\|^2}\right).
\]
For the coarse direction, apply Eq.~\eqref{eq:app-sm-inverse} with $d=r$ and $\mathbf v=\mathbf v_{\!\mathbf B}$ to each row of $\mathcal R_{\!\mathbf B}(\mathbf G)$ to obtain:
\[
\boldsymbol\Sigma_{\!\mathbf B}\mathcal R_{\!\mathbf B}(\mathbf G)
=\frac1\rho\left(\mathbf G\mathbf A^\top-\frac{\mathbf G\mathbf A^\top\mathbf v_{\!\mathbf B}\mathbf v_{\!\mathbf B}^\top}{\rho+\|\mathbf v_{\!\mathbf B}\|^2}\right).
\]
Subtracting the coarse direction from the restricted fine direction cancels the $\mathbf G\mathbf A^\top$ terms and gives:
\[
\boldsymbol\tau_{\mathrm{sgs},\mathbf B}
=\frac1\rho\left[\frac{\mathbf G\mathbf A^\top\mathbf v_{\!\mathbf B}\mathbf v_{\!\mathbf B}^\top}{\rho+\|\mathbf v_{\!\mathbf B}\|^2}
-\frac{\mathbf G\mathbf v_f(\mathbf A\mathbf v_f)^\top}{\rho+\|\mathbf v_f\|^2}\right],
\]
which proves Eq.~\eqref{eq:method-tau-B}.

The same calculation with $\mathcal R_{\!\mathbf A}(\mathbf X)=\mathbf B^\top\mathbf X$ and the A-branch covariance gives:
\[
\boldsymbol\tau_{\mathrm{sgs},\mathbf A}
=\frac1\rho\left[\frac{\mathbf B^\top\mathbf G\mathbf v_{\!\mathbf A}\mathbf v_{\!\mathbf A}^\top}{\rho+\|\mathbf v_{\!\mathbf A}\|^2}
-\frac{\mathbf B^\top\mathbf G\mathbf v_f\mathbf v_f^\top}{\rho+\|\mathbf v_f\|^2}\right],
\]
which proves Eq.~\eqref{eq:method-tau-A}.
\end{proof}

\section{Theoretical Analysis}
\label{sec:app-theoretical-analysis}

This appendix presents the statistical and convergence analysis of ENGINE.
Section~\ref{sec:theory-setup} defines the notation and estimation setting.
Section~\ref{sec:theory-gaussian} derives Gaussian bias and variance bounds.
Section~\ref{sec:theory-mse} compares estimation errors at a common target.
Section~\ref{sec:theory-projection} bounds scalar projection errors.
Section~\ref{sec:theory-oracle} derives the optimal mixing weight.
Section~\ref{sec:theory-convergence} establishes convergence bounds for exact and approximate correction.

\subsection{Setup and assumptions}
\label{sec:theory-setup}

\paragraph{Fine and coarse spaces.}
Subscripts $f$ and $c$ denote fine and coarse spaces, corresponding to full parameters and LoRA factors, respectively.
Write $W=W_0+BA$, with fixed $W_0$, $B\in\mathbb R^{m\times r}$,
and $A\in\mathbb R^{r\times n}$. Using row-wise vectorization, define
$w=\operatorname{vec}(W)\in\mathbb R^{d_f}$ and
$\boldsymbol{\phi}=(\operatorname{vec}(B)^\top,\operatorname{vec}(A)^\top)^\top\in\mathbb R^{d_c}$,
where $d_f=mn$ and $d_c=r(m+n)$.
The fine-space loss is $f(w)$, and its LoRA parameterization is
$F(\boldsymbol{\phi})=f(w(\boldsymbol{\phi}))$.
We omit boldface on matrices and directions in this appendix.

\paragraph{Gradients and restriction.}
The superscript ${\rm raw}$ denotes exact gradients before preconditioning.
We write $g_f^{\rm raw}=\nabla_w f(w)\in\mathbb R^{d_f}$ and
$g_c^{\rm raw}=\nabla_{\boldsymbol{\phi}}F(\boldsymbol{\phi})\in\mathbb R^{d_c}$.
Let $J=\frac{\partial w}{\partial\boldsymbol{\phi}}$ and
$R=J^\top\in\mathbb R^{d_c\times d_f}$ represent prolongation and restriction.
The chain rule gives $g_c^{\rm raw}=R g_f^{\rm raw}$.
For $G\in\mathbb R^{m\times n}$, restriction acts as
$R\operatorname{vec}(G)=(\operatorname{vec}(GA^\top)^\top,\operatorname{vec}(B^\top G)^\top)^\top$.
For the gradient matrix $G=\nabla_W f$, we have $g_f^{\rm raw}=\operatorname{vec}(G)$.

\paragraph{Preconditioned directions and correction.}
Let $\Sigma_f\in\mathbb R^{d_f\times d_f}$ and
$\Sigma_c=\operatorname{diag}(\Sigma_B,\Sigma_A)\in\mathbb R^{d_c\times d_c}$
be symmetric positive-definite preconditioners and perturbation covariances.
The branch blocks have sizes $mr\times mr$ and $rn\times rn$, respectively.
Define the preconditioned directions $g_f=\Sigma_f g_f^{\rm raw}$ and
$g_c=\Sigma_c g_c^{\rm raw}$, the common target $g_\star=R g_f$,
and the exact subgrid residual $\tau=g_\star-g_c$.
The common target $g_\star\in\mathbb R^{d_c}$ applies fine-space preconditioning
before restriction, while $g_c$ applies coarse-space preconditioning after restriction.
Here $g_c$ and $\tau$ are vector forms of $\mathbf s_c$ and
$\boldsymbol\tau_{\mathrm{sgs}}$.

\paragraph{Estimation setting and errors.}
Let $\widehat g_c$ denote the preconditioned coarse estimate and
$\widehat d=\widehat g_c+\lambda\tau$ the corrected direction.
The mean squared error and bias relative to $g_\star$ are
$\mathbb E\|\widehat d-g_\star\|^2$ and $\mathbb E\widehat d-g_\star$, respectively.
For a random vector $X$, $\operatorname{Var}(X)=\mathbb E\|X-\mathbb EX\|^2$ denotes trace variance.
Vector norms are Euclidean.
Expectations condition on current parameters and preconditioners and average over fresh perturbations.
For simplicity, we analyze Gaussian estimators with perturbation radius $\varepsilon$
and $b$ independent directions, where $b$ plays the role of $K$ in the method.
Exact correction uses conditionally deterministic $\tau$ and a fixed mixing weight
$\lambda\in[0,1]$ shared by both branches.
Section~\ref{sec:theory-convergence} specifies the conditions for history-dependent weights.

\subsection{Bias and variance of Gaussian estimators}
\label{sec:theory-gaussian}

Consider an $L$-smooth objective $\varphi$ on $\mathbb R^d$,
$g=\nabla\varphi(x)$, a fixed SPD covariance $\Sigma$ with
$\|\Sigma\|_{\rm op}\le M$, and independent $z_i\sim\mathcal N(0,\Sigma)$.
With deterministic function evaluations, define:
\begin{equation}
\widehat g_{\varepsilon,b}=\frac1b\sum_{i=1}^b
\frac{\varphi(x+\varepsilon z_i)-\varphi(x)}{\varepsilon}z_i,
\qquad \varphi_{\varepsilon,\Sigma}(x)=\mathbb E\varphi(x+\varepsilon z).
\label{eq:theory-rge}
\end{equation}
For fixed $x$, $\Sigma$, and $b$, define the actual trace variance by
$V_\varepsilon=\mathbb E\|\widehat g_{\varepsilon,b}-\mathbb E\widehat g_{\varepsilon,b}\|^2$.
This quantity is defined at every positive smoothing radius.
The subscript $0$ below refers to the estimator obtained in the limit
$\varepsilon\to0$, with the number of random directions $b$ held fixed.
We first derive the mean and bound the smoothing bias separately from sampling variance.

\begin{lemma}[Gaussian smoothing identity]
\label{lem:theory-smoothing}
Under the setting above, the estimator has mean
$\mathbb E\widehat g_{\varepsilon,b}=\Sigma\nabla\varphi_{\varepsilon,\Sigma}(x)$.
Its bias $b_\varepsilon=\mathbb E\widehat g_{\varepsilon,b}-\Sigma g$ obeys
$\|b_\varepsilon\|\le L\varepsilon\|\Sigma\|_{\rm op}
\sqrt{\operatorname{tr}(\Sigma)}\le L\varepsilon\sqrt{dM^3}$.
\end{lemma}
\begin{proof}
Write $z=\Sigma^{\frac{1}{2}}u$ with $u\sim\mathcal N(0,I_d)$.
Since $\mathbb Ez=0$, the constant baseline $\varphi(x)$ contributes
zero to the mean. Gaussian integration by parts, applied componentwise,
gives:
\[
\mathbb E\left[u\varphi(x+\varepsilon\Sigma^{\frac{1}{2}}u)\right]
=\varepsilon\Sigma^{\frac{1}{2}}
\mathbb E\nabla\varphi(x+\varepsilon\Sigma^{\frac{1}{2}}u).
\]
Global $L$-smoothness bounds gradient growth linearly and function growth
quadratically, justifying integration by parts and differentiation under
the Gaussian expectation. Therefore, we have:
\begin{align*}
\mathbb E\widehat g_{\varepsilon,b}
&=\varepsilon^{-1}\Sigma^{\frac{1}{2}}
\mathbb E\left[u\varphi(x+\varepsilon\Sigma^{\frac{1}{2}}u)\right]\\
&=\Sigma\mathbb E\nabla\varphi(x+\varepsilon z)\\
&=\Sigma\nabla\varphi_{\varepsilon,\Sigma}(x).
\end{align*}
Averaging $b$ samples preserves this mean. Subtracting $\Sigma g$ and
using the Lipschitz continuity of the gradient yields:
\begin{align*}
\|b_\varepsilon\|
&=\left\|\Sigma\mathbb E[\nabla\varphi(x+\varepsilon z)-\nabla\varphi(x)]\right\|\\
&\le\|\Sigma\|_{\rm op}\mathbb E
\|\nabla\varphi(x+\varepsilon z)-\nabla\varphi(x)\|\\
&\le L\varepsilon\|\Sigma\|_{\rm op}\mathbb E\|z\|\\
&\le L\varepsilon\|\Sigma\|_{\rm op}\sqrt{\mathbb E\|z\|^2}.
\end{align*}
Using $\mathbb E\|z\|^2=\operatorname{tr}(\Sigma)\le dM$ and
$\|\Sigma\|_{\rm op}\le M$ completes the bias bound.
\end{proof}

The next identity gives the moments of the linear term in the
finite-difference estimator. These moments enter the finite-radius variance bound.

\begin{lemma}[Gaussian covariance identity]
\label{lem:theory-linear-covariance}
For $z\sim\mathcal N(0,\Sigma)$ and fixed $g$, let $Y=(g^\top z)z$. Then we have:
\[
\mathbb EY=\Sigma g,\qquad
\operatorname{Cov}(Y)=(g^\top\Sigma g)\Sigma+\Sigma gg^\top\Sigma.
\]
The covariance of the average of $b$ independent copies is
$\frac{\operatorname{Cov}(Y)}{b}$.
\end{lemma}
\begin{proof}
Since $\mathbb E[zz^\top]=\Sigma$ and $g$ is fixed, the mean is $\mathbb EY=\Sigma g$.
For coordinates $i,j$, expand the second moment and use the Gaussian
fourth-moment identity to obtain:
\begin{align*}
\mathbb E[Y_iY_j]
&=\sum_{k,l}g_kg_l\mathbb E[z_i z_j z_k z_l]\\
&=\sum_{k,l}g_kg_l
(\Sigma_{ij}\Sigma_{kl}+\Sigma_{ik}\Sigma_{jl}+\Sigma_{il}\Sigma_{jk})\\
&=\Sigma_{ij}(g^\top\Sigma g)+2(\Sigma g)_i(\Sigma g)_j.
\end{align*}
Subtracting $(\Sigma g)(\Sigma g)^\top$ gives the stated covariance.
For independent copies $Y^{(1)},\ldots,Y^{(b)}$, cross-covariances vanish,
so $\operatorname{Cov}(b^{-1}\sum_iY^{(i)})=\operatorname{Cov}(Y)/b$.
Taking its trace yields:
\[
\mathbb E\left\|\frac1b\sum_{i=1}^bY^{(i)}-\Sigma g\right\|^2
=\frac{\operatorname{tr}(\Sigma)g^\top\Sigma g+g^\top\Sigma^2g}{b}.
\]
\end{proof}

These identities and a Taylor remainder bound yield the finite-radius estimation error bound.

\begin{theorem}[Gaussian estimation error]
\label{prop:theory-rge}
The mean is $\Sigma\nabla\varphi_{\varepsilon,\Sigma}(x)$ and its bias relative to
$\Sigma g$ satisfies $\|b_\varepsilon\|\le L\varepsilon\sqrt{dM^3}$.
For finite $\varepsilon$, the MSE satisfies:
\begin{equation}
\mathbb E\|\widehat g_{\varepsilon,b}-\Sigma g\|^2
\le \frac{2(d+1)M^2}{b}\|g\|^2
+L^2\varepsilon^2M^3\left[\frac{d(d+2)(d+4)}{2b}+d\right].
\label{eq:theory-finite}
\end{equation}
For fixed $x$, $\Sigma$, and $b$, the estimator converges in mean square
as $\varepsilon\to0$ to
$\widehat g_{0,b}=\frac1b\sum_{i=1}^b(g^\top z_i)z_i$.
\end{theorem}
\begin{proof}
The mean and bias bound follow from Lemma~\ref{lem:theory-smoothing}.
We use Lemma~\ref{lem:theory-linear-covariance} to control the linear term
and bound the remaining finite-difference error below.

\emph{Step 1. Control the Taylor remainder.}
For one sample, let $Y=(g^\top z)z$ and write:
\[
\begin{aligned}
X_\varepsilon&=\frac{\varphi(x+\varepsilon z)-\varphi(x)}{\varepsilon}z\\
&=Y+U_\varepsilon,\\
U_\varepsilon
&=\left(\frac{\varphi(x+\varepsilon z)-\varphi(x)}{\varepsilon}
-g^\top z\right)z.
\end{aligned}
\]
The smoothness inequality gives
$|\varphi(x+\varepsilon z)-\varphi(x)-\varepsilon g^\top z|
\le \frac{L\varepsilon^2\|z\|^2}{2}$, and consequently:
\[
\begin{aligned}
\mathbb E\|U_\varepsilon\|^2&\le\frac{L^2\varepsilon^2}{4}\mathbb E\|z\|^6\\
&\le\frac{L^2\varepsilon^2M^3}{4}d(d+2)(d+4).
\end{aligned}
\]
Here we use $\|\Sigma^{\frac{1}{2}}u\|^2\le M\|u\|^2$ and the Gaussian
moment $\mathbb E\|u\|^6=d(d+2)(d+4)$.
Using the same sampled directions, Jensen's inequality gives:
\[
\begin{aligned}
\mathbb E\|\widehat g_{\varepsilon,b}-\widehat g_{0,b}\|^2&=\mathbb E\left\|\frac1b\sum_{i=1}^bU_{\varepsilon,i}\right\|^2\\
&\le\frac1b\sum_{i=1}^b\mathbb E\|U_{\varepsilon,i}\|^2
\longrightarrow0
\quad\text{as }\varepsilon\to0.
\end{aligned}
\]
For fixed $b$, this proves mean-square convergence to $\widehat g_{0,b}$
and $V_\varepsilon\to V_0$ as $\varepsilon\to0$.

\emph{Step 2. Bound the centered variance.}
Applying $\|a+b\|^2\le2\|a\|^2+2\|b\|^2$ to the centered
$Y$ and $U_\varepsilon$ gives:
\begin{align*}
\operatorname{Var}(X_\varepsilon)
&\le2\operatorname{Var}(Y)+2\operatorname{Var}(U_\varepsilon)\\
&\le2[\operatorname{tr}(\Sigma)g^\top\Sigma g+g^\top\Sigma^2g]
+2\mathbb E\|U_\varepsilon\|^2\\
&\le2(d+1)M^2\|g\|^2
+\frac{L^2\varepsilon^2M^3}{2}d(d+2)(d+4).
\end{align*}
Here $g^\top\Sigma g\le M\|g\|^2$,
$g^\top\Sigma^2g\le M^2\|g\|^2$, and
$\operatorname{tr}(\Sigma)\le dM$. Averaging $b$ independent
$X_\varepsilon$ samples divides this variance bound by $b$.

\emph{Step 3. Add the squared smoothing bias.}
Write $\widehat g_{\varepsilon,b}-\Sigma g
=(\widehat g_{\varepsilon,b}-\mathbb E\widehat g_{\varepsilon,b})+b_\varepsilon$.
The centered term has zero mean, so:
\begin{align*}
\mathbb E\|\widehat g_{\varepsilon,b}-\Sigma g\|^2
&=\operatorname{Var}(\widehat g_{\varepsilon,b})+\|b_\varepsilon\|^2\\
&\le\frac{2(d+1)M^2}{b}\|g\|^2
+\frac{L^2\varepsilon^2M^3}{2b}d(d+2)(d+4)
+L^2\varepsilon^2dM^3.
\end{align*}
This is \eqref{eq:theory-finite}. The variance decreases with $b$,
while the smoothing-bias term is controlled by $\varepsilon$.
\end{proof}

\paragraph{Local-linear limit.}
By Theorem~\ref{prop:theory-rge}, $b_\varepsilon\to0$ and
$\widehat g_{\varepsilon,b}\to\widehat g_{0,b}$ in mean square as
$\varepsilon\to0$ with $b$ fixed. Applying
Lemma~\ref{lem:theory-linear-covariance} to this limiting estimator gives
$\mathbb E\widehat g_{0,b}=\Sigma g$. Its covariance and trace variance are:
\begin{equation}
\begin{aligned}
\operatorname{Cov}(\widehat g_{0,b})
&=\frac{(g^\top\Sigma g)\Sigma+\Sigma gg^\top\Sigma}{b},\\
V_0&=\frac{\operatorname{tr}(\Sigma)g^\top\Sigma g+g^\top\Sigma^2g}{b}.
\end{aligned}
\label{eq:theory-variance}
\end{equation}
The limiting MSE relative to $\Sigma g$ is therefore $V_0$.
For $\Sigma=I_d$, substitution gives
$V_0=\frac{(d+1)\|g\|^2}{b}$.
All constants here concern the Gaussian estimator in Eq.~\eqref{eq:theory-rge}.

\subsection{Common-target MSE and exact correction}
\label{sec:theory-mse}

Section~\ref{sec:theory-finite-comparison} treats finite perturbations
with exact correction, and Section~\ref{sec:theory-approximate-comparison}
includes correction error. Section~\ref{sec:theory-local-comparison}
then specializes these results to the local-linear limit.

\paragraph{Which estimators and which targets?}
We distinguish isotropic LoRA ZO from LoRA+LOREN. The former samples
with covariance $I_{d_c}$ and targets $g_c^{\rm raw}$, whereas the latter samples
with covariance $\Sigma_c$ and targets $g_c=\Sigma_c g_c^{\rm raw}$.
Full-space LOREN samples with covariance $\Sigma_f$ and targets
$g_f=\Sigma_f g_f^{\rm raw}$. Ideal ENGINE uses the same coarse
estimator as LoRA+LOREN, shifted by the exact residual $\tau$.
The comparison uses the Gaussian estimator models of
\eqref{eq:theory-rge}, which isolate the effects of sampling covariance
and subgrid correction.
For $s\in\{f,c\}$, superscripts ${\rm iso}$ and ${\rm ani}$ denote
sampling covariances $I_{d_s}$ and $\Sigma_s$, respectively.
We abbreviate the Gaussian coarse estimator as
$\widehat g_c=\widehat g_c^{\rm ani}$, suppressing $\varepsilon$ and $b$.
This is $\widehat{\mathbf g}_{c,\epsilon,b}$ from
Eq.~\eqref{eq:method-gaussian-direction}.

\subsubsection{Finite-radius analysis}
\label{sec:theory-finite-comparison}

We first keep $\varepsilon>0$. Let $f$ and $F$ have smoothness constants
$L_f,L_c$, and let $\|\Sigma_s\|_{\rm op}\le M_s$ for $s\in\{f,c\}$.
Theorem~\ref{prop:theory-rge} gives the variance and smoothing-bias bounds:
\begin{align}
\operatorname{Var}(\widehat g_s^{\rm ani})
&\le\frac{2(d_s+1)M_s^2}{b}\|g_s^{\rm raw}\|^2
+\frac{L_s^2\varepsilon^2M_s^3}{2b}d_s(d_s+2)(d_s+4),
\label{eq:theory-comparison-variance-bound}\\
\|\mathbb E\widehat g_s^{\rm ani}-g_s\|
&\le L_s\varepsilon\sqrt{d_sM_s^3}.
\label{eq:theory-comparison-bias-bound}
\end{align}
For isotropic sampling, set $\Sigma_s=I_{d_s}$ and $M_s=1$, with target
$g_s^{\rm raw}$. All comparisons use the same direction count $b$ and radius
$\varepsilon$.

For any deterministic target $a$, expanding the centered error gives:
\begin{equation}
\mathbb E\|X-a\|^2
=\operatorname{Var}(X)+\|\mathbb EX-a\|^2,
\label{eq:theory-bias-variance-identity}
\end{equation}
since $\mathbb E(X-\mathbb EX)=0$ makes the cross term vanish.
For isotropic full-space and coarse estimators, use targets
$g_f^{\rm raw}$ and $g_c^{\rm raw}$, respectively. Full-space anisotropic
sampling targets $g_f$. To compare coarse updates with and without
correction, we use the common target $g_\star$ below.

For fixed $b$ and $\varepsilon>0$, the variance bound contains a term
linear in $d_s$ and a remainder of order $\varepsilon^2d_s^3$.
With comparable gradient, covariance, and smoothness scales bounded
independently of dimension, coarse sampling reduces these dimension
factors from $d_f=mn$ to $d_c=r(m+n)$ when $d_c\ll d_f$.

\paragraph{Exact correction at a common target.}
We next quantify how exact correction changes the bias and MSE while
preserving the coarse sampling variance. Let $\widehat g_c$ denote the
finite-radius coarse estimator, with smoothing bias
$b_{c,\varepsilon}=\mathbb E\widehat g_c-g_c$ and trace variance
$V_c=\operatorname{Var}(\widehat g_c)$. For fixed $\lambda\in[0,1]$,
the corrected direction is $\widehat d_\lambda=\widehat g_c+\lambda\tau$.

\begin{proposition}[Exact-correction benefit]
\label{prop:theory-mse}
The ideal estimator satisfies:
\begin{equation}
\operatorname{Var}(\widehat d_\lambda)=V_c,\qquad
\mathbb E\|\widehat d_\lambda-g_\star\|^2
=V_c+\|b_{c,\varepsilon}-(1-\lambda)\tau\|^2.
\label{eq:theory-mse}
\end{equation}
The MSE reduction relative to the uncorrected estimator is:
\[
\mathbb E\|\widehat g_c-g_\star\|^2
-\mathbb E\|\widehat d_\lambda-g_\star\|^2
=\lambda(2-\lambda)\|\tau\|^2
-2\lambda\langle b_{c,\varepsilon},\tau\rangle.
\]
\end{proposition}
\begin{proof}
Conditional on the state, $\tau$ and $\lambda$ are deterministic, so:
\[
\begin{aligned}
\operatorname{Var}(\widehat d_\lambda)&=\operatorname{Var}(\widehat g_c),\\
\mathbb E\widehat d_\lambda-g_\star
&=b_{c,\varepsilon}-(1-\lambda)\tau.
\end{aligned}
\]
Applying the bias-variance identity in
Eq.~\eqref{eq:theory-bias-variance-identity} proves Eq.~\eqref{eq:theory-mse}.
Subtracting the result at $\lambda$ from that at $\lambda=0$ gives:
\[
\|b_{c,\varepsilon}-\tau\|^2
-\|b_{c,\varepsilon}-(1-\lambda)\tau\|^2
=\lambda(2-\lambda)\|\tau\|^2
-2\lambda\langle b_{c,\varepsilon},\tau\rangle.
\]
\end{proof}

For finite $\varepsilon$ and full correction $\lambda=1$, the same identity gives:
\begin{align*}
\operatorname{MSE}_{\rm LoRA+LOREN}-\operatorname{MSE}_{\rm ideal}
&=\|\tau\|^2-2\langle b_{c,\varepsilon},\tau\rangle\\
&\ge\|\tau\|(\|\tau\|-2L_c\varepsilon\sqrt{d_cM_c^3}).
\end{align*}
The inequality follows from Cauchy-Schwarz and
Eq.~\eqref{eq:theory-comparison-bias-bound}. Thus
$\|\tau\|>2L_c\varepsilon\sqrt{d_cM_c^3}$ is sufficient for strictly
smaller squared bias and MSE, with unchanged coarse sampling variance.

\subsubsection{Approximate correction}
\label{sec:theory-approximate-comparison}

At finite $\varepsilon$, an approximate correction can add bias and
sampling noise. The following decomposition includes their dependence
on the coarse estimate and provides the objective for choosing the mixing weight.
Here $\widehat g_c$ denotes the finite-radius coarse estimator,
with trace variance $V_c$, smoothing bias
$b_{c,\varepsilon}$, and centered fluctuation $\xi_c=\widehat g_c-\mathbb E\widehat g_c$.

For a general correction $\widehat\tau=\tau+b_\tau+\xi_\tau$,
define $b_\tau=\mathbb E\widehat\tau-\tau$ and
$\xi_\tau=\widehat\tau-\mathbb E\widehat\tau$, so $\mathbb E\xi_\tau=0$.
Let $V_\tau=\mathbb E\|\xi_\tau\|^2$ and
$C_{c,\tau}=\mathbb E\langle\xi_c,\xi_\tau\rangle$.
\begin{proposition}[MSE with an approximate correction]
\label{prop:theory-noisy-mse}
For fixed $\lambda\in[0,1]$ and square-integrable estimates, the MSE is:
\begin{equation}
\begin{aligned}
\mathcal E(\lambda)&:=\mathbb E\|\widehat g_c+\lambda\widehat\tau-g_\star\|^2\\
&=V_c+\lambda^2V_\tau+2\lambda C_{c,\tau}
+\|b_{c,\varepsilon}-(1-\lambda)\tau+\lambda b_\tau\|^2.
\end{aligned}
\label{eq:theory-general-mse}
\end{equation}
\end{proposition}
\begin{proof}
The error splits into a centered fluctuation and a deterministic bias:
\[
\widehat g_c+\lambda\widehat\tau-g_\star
=\xi_c+\lambda\xi_\tau
+b_{c,\varepsilon}-(1-\lambda)\tau+\lambda b_\tau.
\]
The first two terms sum to a zero-mean vector. Their squared norm has
expectation:
\begin{align*}
\mathbb E\|\xi_c+\lambda\xi_\tau\|^2
&=\mathbb E\|\xi_c\|^2+\lambda^2\mathbb E\|\xi_\tau\|^2
+2\lambda\mathbb E\langle\xi_c,\xi_\tau\rangle\\
&=V_c+\lambda^2V_\tau+2\lambda C_{c,\tau}.
\end{align*}
The expected cross term vanishes, so adding the squared bias yields \eqref{eq:theory-general-mse}.
\end{proof}
The covariance term $C_{c,\tau}$ accounts for dependence between the two
estimates. For history-measurable weights, the formula applies
conditional on that history. For weights computed from the same samples,
the analysis applies directly to the joint quantity $\lambda\widehat\tau$.

\subsubsection{Local-linear limit and exact comparisons}
\label{sec:theory-local-comparison}

All estimators and variances in this subsection refer to the limit
$\varepsilon\to0$ with $b$ fixed.

\begin{proof}[Proof of Proposition~\ref{prop:method-common-target}]
Apply Theorem~\ref{prop:theory-rge} with $\varphi=F$ and $\Sigma=\Sigma_c$
to obtain mean-square convergence and $b_{c,\varepsilon}\to0$.
Lemma~\ref{lem:theory-linear-covariance} gives the limiting mean $g_c$
and trace variance $V_{c,0}$ in Eq.~\eqref{eq:method-local-variance}.
Setting $b_{c,\varepsilon}=0$ in Proposition~\ref{prop:theory-mse} gives:
\[
\mathbb E\|\widehat d_\lambda-g_\star\|^2
=V_c+(1-\lambda)^2\|\tau\|^2.
\]
Here $V_c$ denotes $V_{c,0}$ in the main text.
Setting $\lambda=0$ gives Eq.~\eqref{eq:method-coarse-mse}, and
$\lambda=1$ gives Eq.~\eqref{eq:method-ideal-mse}.
\end{proof}
More generally, the MSE reduction is
$\lambda(2-\lambda)\|\tau\|^2$, which is positive for
$0<\lambda\le1$ and $\tau\ne0$.
For approximate corrections with mean-square limits, substituting
$b_{c,\varepsilon}=0$ and the limiting correction moments into
Proposition~\ref{prop:theory-noisy-mse} gives the corresponding local result.

\paragraph{Dimension dependence of sampling variance.}
For fixed $b$ and comparable gradient and covariance scales bounded
independently of dimension, the local-linear variance bounds give
$V_f^{\rm ani}=O(d_f)$ and $V_c^{\rm ani}=O(d_c)$.
Here $V_f^{\rm ani}$ is the variance measured in full parameter coordinates.
The linear dimension factor explains the sampling advantage of
$d_c\ll d_f$. We next restrict the full-space output to compare all
estimators in the same coarse coordinates.

\paragraph{Bias and variance at a common target.}
To compare full-space LOREN with coarse estimators, first restrict its
output to the same coordinates using $R\widehat g_f^{\rm ani}$.
Define $\delta_{\rm iso}=g_\star-g_c^{\rm raw}$.
Let $V_c^{\rm iso}$ and $V_c^{\rm ani}$ denote the trace variances of
$\widehat g_{c,0,b}^{\rm iso}$ and $\widehat g_{c,0,b}^{\rm ani}$, respectively,
and let $V_\star^{\rm direct}$ denote that of $R\widehat g_{f,0,b}^{\rm ani}$.
Substituting the coarse gradients and covariances into
Eq.~\eqref{eq:theory-variance} gives the first two expressions below.
For the third, apply the same identity with the fine-space quantities
and use $\operatorname{Cov}(RX)=R\operatorname{Cov}(X)R^\top$.
Their exact expressions are:
\begin{align}
V_c^{\rm iso}&=\frac{(d_c+1)\|g_c^{\rm raw}\|^2}{b},\\
V_c^{\rm ani}&=\frac{\operatorname{tr}(\Sigma_c)(g_c^{\rm raw})^\top\Sigma_c g_c^{\rm raw}
+(g_c^{\rm raw})^\top\Sigma_c^2g_c^{\rm raw}}{b},\\
V_\star^{\rm direct}
&=\frac{((g_f^{\rm raw})^\top\Sigma_f g_f^{\rm raw})
\operatorname{tr}(R\Sigma_fR^\top)+\|g_\star\|^2}{b}.
\label{eq:theory-restricted-variance}
\end{align}
Here $R\Sigma_f g_f^{\rm raw}=g_\star$, and taking the trace of the
restricted covariance gives $V_\star^{\rm direct}$.

\begin{table}[htbp]
\centering
\caption{Exact comparison in the local-linear limit $\varepsilon\to0$,
with $b$ held fixed, against the
\emph{same} target $g_\star$, conditional on the current state.
Bias is mean minus target. All methods use $b$ independent directions.
\Engine\ rows use exact, conditionally deterministic correction.}
\label{tab:theory-common-target}
\small
\setlength{\tabcolsep}{3pt}
\begin{tabular}{llll}
\toprule
Estimator & Bias & Trace variance & MSE \\
\midrule
Isotropic LoRA RGE & $-\delta_{\rm iso}$ & $V_c^{\rm iso}$
& $V_c^{\rm iso}+\|\delta_{\rm iso}\|^2$ \\
LoRA+LOREN & $-\tau$ & $V_c^{\rm ani}$
& $V_c^{\rm ani}+\|\tau\|^2$ \\
Restricted full-space LOREN & $0$ & $V_\star^{\rm direct}$
& $V_\star^{\rm direct}$ \\
Ideal \Engine\ ($\lambda\in[0,1]$) & $-(1-\lambda)\tau$ & $V_c^{\rm ani}$
& $V_c^{\rm ani}+(1-\lambda)^2\|\tau\|^2$ \\
Ideal \Engine\ ($\lambda=1$) & $0$ & $V_c^{\rm ani}$
& $V_c^{\rm ani}$ \\
\bottomrule
\end{tabular}
\end{table}

Table~\ref{tab:theory-common-target} follows by inserting the respective
means $g_c^{\rm raw}$, $g_c$, $g_\star$, and $g_c+\lambda\tau$ into
\eqref{eq:theory-bias-variance-identity}. Ideal ENGINE with $\lambda=1$
has smaller MSE than
restricted full-space LOREN precisely when
$V_c^{\rm ani}<V_\star^{\rm direct}$, and smaller MSE than isotropic
LoRA precisely when
$V_c^{\rm ani}<V_c^{\rm iso}+\|\delta_{\rm iso}\|^2$.
These conditions account for gradients, covariances, and restriction in each comparison.
\paragraph{Combined benefit of ideal ENGINE.}
The two benefits act on separate terms of the MSE decomposition.
Coarse sampling reduces the dimension-dependent variance under the stated
scale assumptions. Exact correction recovers $g_\star$ and removes the
structural bias, reducing $V_c^{\rm ani}+\|\tau\|^2$ to $V_c^{\rm ani}$.

\subsection{Scalar approximation error}
\label{sec:theory-projection}

The exact branch residual has the form
$\tau_\ell=\rho^{-1}(\mathbf c_\ell q_\ell^\top-\mathbf c_{f,\ell}s_\ell^\top)$,
where $p_B=m$, $p_A=r$, $\mathbf c_\ell,\mathbf c_{f,\ell}\in\mathbb R^{p_\ell}$,
and $\rho>0$ is the same damping parameter as in the rank-1 preconditioner.
Here $v_B$, $v_A$, and $v_f$ denote the method's curvature vectors
$\mathbf v_{\!\mathbf B}$, $\mathbf v_{\!\mathbf A}$, and $\mathbf v_f$,
respectively. The branch residual $\tau_\ell$ is the matrix form of
$\boldsymbol\tau_{\mathrm{sgs},\ell}$, for $\ell\in\{B,A\}$.
Specifically:
\[
\begin{array}{ll}
q_B=v_B,\quad s_B=Av_f,&q_A=v_A,\quad s_A=v_f,\\[6pt]
\displaystyle \mathbf c_B=\frac{GA^\top v_B}{\rho+\|v_B\|^2},&
\displaystyle \mathbf c_{f,B}=\frac{Gv_f}{\rho+\|v_f\|^2},\\[6pt]
\displaystyle \mathbf c_A=\frac{B^\top Gv_A}{\rho+\|v_A\|^2},&
\displaystyle \mathbf c_{f,A}=\frac{B^\top Gv_f}{\rho+\|v_f\|^2}.
\end{array}
\]
These are the exact coefficients of
Proposition~\ref{prop:subgrid-residual}. Their row averages define the
scalar-broadcast approximation studied below.
Let $\mathbf1_p$ be the all-ones vector,
$\Pi_p=\frac{\mathbf1_p\mathbf1_p^\top}{p}$ and $P_p=I_p-\Pi_p$.
For $p=p_\ell$, define the arithmetic coefficient means
$\bar c_\ell=\frac{\mathbf1_p^\top \mathbf c_\ell}{p}$ and
$\bar c_{f,\ell}=\frac{\mathbf1_p^\top \mathbf c_{f,\ell}}{p}$.

The next proposition separates row compression error from scalar coefficient mismatch.

\begin{proposition}[Row compression and scalar approximation error]
\label{prop:theory-projection}
For $p=p_\ell$, the row mean $\tau_\ell^{\rm mean}=\Pi_p\tau_\ell$
is the unique closest matrix to $\tau_\ell$ among matrices with identical
rows, in the Frobenius norm. Let $D_\ell=\|P_p\tau_\ell\|_F$.
For any matrix $Z$ with identical rows, orthogonality gives:
\begin{equation}
\|\tau_\ell-Z\|_F^2=D_\ell^2+\|\tau_\ell^{\rm mean}-Z\|_F^2.
\label{eq:theory-projection}
\end{equation}
In particular, for scalar coefficients $\widetilde c_\ell,\widetilde c_{f,\ell}$,
set $\widetilde\tau_\ell=\frac1\rho\mathbf1_p
(\widetilde c_\ell q_\ell^\top-\widetilde c_{f,\ell}s_\ell^\top)$. Then we have:
\begin{equation}
\|\tau_\ell-\widetilde\tau_\ell\|_F^2
=D_\ell^2+\frac p{\rho^2}
\|(\widetilde c_\ell-\bar c_\ell)q_\ell
-(\widetilde c_{f,\ell}-\bar c_{f,\ell})s_\ell\|^2.
\label{eq:theory-scalar-extra}
\end{equation}
\end{proposition}
\begin{proof}
Row averaging gives
$\tau_\ell^{\rm mean}=\frac1\rho\mathbf1_p
(\bar c_\ell q_\ell^\top-\bar c_{f,\ell}s_\ell^\top)$.
The rows of $P_p\tau_\ell$ sum to zero, so this residual is orthogonal
to every matrix with identical rows. Therefore, we have:
\begin{align*}
\|\tau_\ell-Z\|_F^2
&=\|P_p\tau_\ell+(\tau_\ell^{\rm mean}-Z)\|_F^2\\
&=\|P_p\tau_\ell\|_F^2+\|\tau_\ell^{\rm mean}-Z\|_F^2.
\end{align*}
The second term vanishes exactly at $Z=\tau_\ell^{\rm mean}$, proving
optimality and uniqueness. For $Z=\widetilde\tau_\ell$, its value is:
\[
\|\widetilde\tau_\ell-\tau_\ell^{\rm mean}\|_F^2
=\frac p{\rho^2}
\|(\widetilde c_\ell-\bar c_\ell)q_\ell
-(\widetilde c_{f,\ell}-\bar c_{f,\ell})s_\ell\|^2,
\]
since $\|\mathbf1_p\|^2=p$. Substitution proves
Eq.~\eqref{eq:theory-scalar-extra}.
\end{proof}

The two branches have minimum total compression error $D_B^2+D_A^2$.
For general scalar coefficients, the total squared correction error
is the sum of both right-hand sides of Eq.~\eqref{eq:theory-scalar-extra}.
Appendix~\ref{sec:app-surrogate} specifies how the implementation's
normalization and scaling determine these coefficients.

\subsection{Oracle-optimal mixing}
\label{sec:theory-oracle}

The MSE is quadratic in $\lambda$. Define $\bar\tau=\mathbb E\widehat\tau=\tau+b_\tau$.
\begin{proposition}[Oracle-optimal mixing weight]
\label{prop:theory-oracle}
If $\|\bar\tau\|^2+V_\tau>0$, minimizing
\eqref{eq:theory-general-mse} over fixed $\lambda\in[0,1]$ gives:
\begin{equation}
\lambda^\star=\operatorname{clip}_{[0,1]}
\left(\frac{\langle\tau-b_{c,\varepsilon},\bar\tau\rangle-C_{c,\tau}}
{\|\bar\tau\|^2+V_\tau}\right).
\end{equation}
If the denominator vanishes, $\widehat\tau=0$ almost surely and every
feasible weight is optimal.
\end{proposition}
\begin{proof}
Set $a_\tau=\|\bar\tau\|^2+V_\tau$ and
$q_\tau=\langle\tau-b_{c,\varepsilon},\bar\tau\rangle-C_{c,\tau}$.
Expanding the squared bias in Eq.~\eqref{eq:theory-general-mse} gives:
\[
\mathcal E(\lambda)=\mathcal E(0)+a_\tau\lambda^2-2q_\tau\lambda.
\]
For $a_\tau>0$, this strictly convex quadratic is minimized at
$q_\tau/a_\tau$ on $\mathbb R$, and clipping gives the minimum on $[0,1]$.
The same argument applies to any closed subinterval
$[\lambda_{\min},\lambda_{\max}]$.
If $a_\tau=\mathbb E\|\widehat\tau\|^2=0$, then
$\widehat\tau=0$ almost surely and $\mathcal E$ is independent of $\lambda$.
\end{proof}

For unbiased estimators with uncorrelated noise, set
$b_{c,\varepsilon}=0$, $b_\tau=0$, and $C_{c,\tau}=0$.
When $\|\tau\|^2+V_\tau>0$, substituting these conditions into
Proposition~\ref{prop:theory-oracle} and Eq.~\eqref{eq:theory-general-mse} gives:
\[
\lambda^\star=\frac{\|\tau\|^2}{\|\tau\|^2+V_\tau},
\qquad
\mathcal E(\lambda^\star)=
V_c+\frac{\|\tau\|^2V_\tau}{\|\tau\|^2+V_\tau}.
\]
For fixed positive correction energy, this optimal weight decreases as the
correction variance grows.

\subsection{Convergence under positive alignment}
\label{sec:theory-convergence}

We analyze updates on the fixed coarse objective
$F(\boldsymbol\phi)=f(w(\boldsymbol\phi))$ with $W_0$ held fixed.
Stationarity is measured by the gradient with respect to the stacked LoRA
factors, $g_{c,t}^{\rm raw}=\nabla_{\boldsymbol\phi}F(\boldsymbol\phi_t)$.
Let $\mathcal F_t$ be the history before fresh step-$t$ perturbations,
and $\mathbb E_t=\mathbb E[\cdot\mid\mathcal F_t]$.
All current parameters, preconditioners, and exact targets are
$\mathcal F_t$-measurable.

\paragraph{Assumptions for ENGINE convergence.}
The following assumptions hold uniformly over the coarse steps under analysis.
\begin{enumerate}[label=(A\arabic*),ref=A\arabic*,leftmargin=*,itemsep=3pt]
\item \textbf{Smoothness and lower boundedness.}\label{ass:theory-smoothness}
The coarse objective $F$ is $L_c$-smooth with $L_c>0$ and
$F(\boldsymbol{\phi})\ge F_{\inf}$.
\item \textbf{Bounded positive-definite preconditioning.}\label{ass:theory-preconditioner}
The coarse preconditioner satisfies
$m_cI\preceq\Sigma_{c,t}\preceq M_cI$ for constants $0<m_c\le M_c$.
\item \textbf{Target alignment and norm control.}\label{ass:theory-alignment}
The restricted full-space target satisfies:
\begin{equation}
\langle g_{c,t}^{\rm raw},g_{\star,t}\rangle\ge\kappa_\star\|g_{c,t}^{\rm raw}\|^2,
\qquad\|g_{\star,t}\|\le M_\star\|g_{c,t}^{\rm raw}\|,
\qquad\kappa_\star>0.
\label{eq:theory-alignment}
\end{equation}
\item \textbf{Predictable mixing.}\label{ass:theory-mixing}
The weight $\lambda_t\in[0,1]$ is $\mathcal F_t$-measurable, so it is
fixed before the fresh step-$t$ perturbations are sampled.
\item \textbf{Controlled estimator bias and variance.}\label{ass:theory-estimator}
For nonnegative constants $\beta_c,\nu_c,\omega_c$, the coarse estimator satisfies:
\[
\|\mathbb E_t\widehat g_{c,t}-g_{c,t}\|\le\beta_c,\qquad
\operatorname{Var}_t(\widehat g_{c,t})\le\nu_c\|g_{c,t}^{\rm raw}\|^2+\omega_c.
\]
\end{enumerate}

\paragraph{Empirical alignment proxy.}
In our diagnostic runs, the cosine between the raw gradient estimate and
the deployed pre-momentum direction after correction and clipping was
positive at every measured coarse step. This measures alignment with the
estimated gradient, while Assumption~\ref{ass:theory-alignment} concerns
uniform alignment and norm control of the exact target.

The following descent lemma provides a common bound for exact and
approximate correction.

\begin{lemma}[Conditional descent with a biased stochastic direction]
\label{lem:theory-descent}
Suppose $F$ is $L_c$-smooth, $L_c>0$, and
$\boldsymbol{\phi}_{t+1}=\boldsymbol{\phi}_t-\eta\widehat d_t$.
Let an $\mathcal F_t$-measurable reference direction $d_t$ satisfy:
\[
\begin{aligned}
\langle g_{c,t}^{\rm raw},d_t\rangle&\ge\kappa\|g_{c,t}^{\rm raw}\|^2,\\
\|d_t\|&\le M\|g_{c,t}^{\rm raw}\|,\qquad \kappa>0.
\end{aligned}
\]
Assume
$\|\mathbb E_t\widehat d_t-d_t\|\le\beta$ and
$\operatorname{Var}_t(\widehat d_t)\le\nu\|g_{c,t}^{\rm raw}\|^2+\omega$,
where $\beta,\nu,\omega$ are nonnegative constants.
Set $\mathcal A=2M^2+\nu>0$ and $\mathcal B=2\beta^2+\omega$.
For $0<\eta\le\frac{\kappa}{2L_c\mathcal A}$, the one-step bound is:
\begin{equation}
\mathbb E_tF(\boldsymbol{\phi}_{t+1})
\le F(\boldsymbol{\phi}_t)-\frac{\eta\kappa}{4}\|g_{c,t}^{\rm raw}\|^2
+\frac{\eta\beta^2}{2\kappa}+\frac{L_c\eta^2\mathcal B}{2}.
\label{eq:theory-one-step}
\end{equation}
\end{lemma}
\begin{proof}
\emph{Step 1. Lower-bound the mean descent term.}
Let $e_t=\mathbb E_t\widehat d_t-d_t$.
The Cauchy-Schwarz inequality and $\|e_t\|\le\beta$ imply:
\[
\begin{aligned}
\langle g_{c,t}^{\rm raw},\mathbb E_t\widehat d_t\rangle&=\langle g_{c,t}^{\rm raw},d_t\rangle+\langle g_{c,t}^{\rm raw},e_t\rangle\\
&\ge\kappa\|g_{c,t}^{\rm raw}\|^2-\beta\|g_{c,t}^{\rm raw}\|.
\end{aligned}
\]
Young's inequality in the form
$\beta\|g_{c,t}^{\rm raw}\|\le\frac{\kappa\|g_{c,t}^{\rm raw}\|^2}{2}+\frac{\beta^2}{2\kappa}$ gives:
\[
\langle g_{c,t}^{\rm raw},\mathbb E_t\widehat d_t\rangle
\ge\frac\kappa2\|g_{c,t}^{\rm raw}\|^2-\frac{\beta^2}{2\kappa}.
\]

\emph{Step 2. Upper-bound the second moment.}
Separating conditional variance from the squared conditional mean yields:
\begin{align*}
\mathbb E_t\|\widehat d_t\|^2
&=\|\mathbb E_t\widehat d_t\|^2+\operatorname{Var}_t(\widehat d_t)\\
&=\|d_t+e_t\|^2+\operatorname{Var}_t(\widehat d_t)\\
&\le2\|d_t\|^2+2\|e_t\|^2+\nu\|g_{c,t}^{\rm raw}\|^2+\omega\\
&\le(2M^2+\nu)\|g_{c,t}^{\rm raw}\|^2+2\beta^2+\omega\\
&=\mathcal A\|g_{c,t}^{\rm raw}\|^2+\mathcal B.
\end{align*}

\emph{Step 3. Apply smoothness and the step-size condition.}
The update and $L_c$-smoothness imply:
\begin{align*}
\mathbb E_tF(\boldsymbol{\phi}_{t+1})
&\le F(\boldsymbol{\phi}_t)-\eta\langle g_{c,t}^{\rm raw},\mathbb E_t\widehat d_t\rangle
+\frac{L_c\eta^2}{2}\mathbb E_t\|\widehat d_t\|^2\\
&\le F(\boldsymbol{\phi}_t)
-\left(\frac{\eta\kappa}{2}-\frac{L_c\eta^2\mathcal A}{2}\right)\|g_{c,t}^{\rm raw}\|^2
+\frac{\eta\beta^2}{2\kappa}+\frac{L_c\eta^2\mathcal B}{2}.
\end{align*}
Since $\frac{L_c\eta^2\mathcal A}{2}\le\frac{\eta\kappa}{4}$, the coefficient
of the squared gradient is at least $\frac{\eta\kappa}{4}$.
This proves \eqref{eq:theory-one-step}.
\end{proof}

For ideal ENGINE, use the exact correction
$\widehat d_t=\widehat g_{c,t}+\lambda_t\tau_t$.
Set $\kappa=\min\{m_c,\kappa_\star\}$,
$M=\max\{M_c,M_\star\}$,
$\mathcal A=2M^2+\nu_c$ and $\mathcal B=2\beta_c^2+\omega_c$.

Let $T\ge1$ be the number of optimization steps.
The next theorem bounds the average squared coarse gradient after these
steps and identifies the residual error due to estimator bias.

\begin{theorem}[Ideal ENGINE stationarity]
\label{prop:theory-convergence}
Under Assumptions~\ref{ass:theory-smoothness} to \ref{ass:theory-estimator},
for $\boldsymbol{\phi}_{t+1}=\boldsymbol{\phi}_t-\eta\widehat d_t$,
$\eta=\frac{\kappa}{2L_c\mathcal A\sqrt{T}}$ and $T\ge1$, the iterates satisfy:
\begin{equation}
\frac{1}{T}\sum_{t=0}^{T-1}\mathbb E\|g_{c,t}^{\rm raw}\|^2
\le\frac{8L_c\mathcal A\Delta_F}{\kappa^2\sqrt{T}}
+\frac{\mathcal B}{\mathcal A\sqrt{T}}
+\frac{2\beta_c^2}{\kappa^2},
\qquad\Delta_F=F(\boldsymbol{\phi}_0)-F_{\inf}.
\label{eq:theory-rate}
\end{equation}
\end{theorem}
\begin{proof}
\emph{Step 1. Verify alignment and the norm bound for the mixed target.}
The mixed target satisfies:
\[
\begin{aligned}
d_t&=g_{c,t}+\lambda_t\tau_t\\
&=(1-\lambda_t)g_{c,t}+\lambda_tg_{\star,t}.
\end{aligned}
\]
Since $g_{c,t}=\Sigma_{c,t}g_{c,t}^{\rm raw}$, the spectral lower bound and
\eqref{eq:theory-alignment} give:
\begin{align*}
\langle g_{c,t}^{\rm raw},d_t\rangle
&=(1-\lambda_t)(g_{c,t}^{\rm raw})^\top\Sigma_{c,t}g_{c,t}^{\rm raw}
+\lambda_t\langle g_{c,t}^{\rm raw},g_{\star,t}\rangle\\
&\ge[(1-\lambda_t)m_c+\lambda_t\kappa_\star]\|g_{c,t}^{\rm raw}\|^2\\
&\ge\kappa\|g_{c,t}^{\rm raw}\|^2.
\end{align*}
Similarly, the triangle inequality and the upper bounds yield:
\[
\begin{aligned}
\|d_t\|&\le(1-\lambda_t)\|\Sigma_{c,t}g_{c,t}^{\rm raw}\|
+\lambda_t\|g_{\star,t}\|\\
&\le[(1-\lambda_t)M_c+\lambda_tM_\star]\|g_{c,t}^{\rm raw}\|\\
&\le M\|g_{c,t}^{\rm raw}\|.
\end{aligned}
\]
Both comparisons use $0\le\lambda_t\le1$.

\emph{Step 2. Apply the exact-correction result conditionally.}
Predictability makes $\lambda_t$ fixed conditional on $\mathcal F_t$.
The same holds for $\tau_t$, so Proposition~\ref{prop:theory-mse} gives:
\[
\begin{aligned}
\mathbb E_t\widehat d_t-d_t&=\mathbb E_t\widehat g_{c,t}-g_{c,t},\\
\operatorname{Var}_t(\widehat d_t)&=\operatorname{Var}_t(\widehat g_{c,t}).
\end{aligned}
\]
Thus Lemma~\ref{lem:theory-descent} applies with
$\beta=\beta_c$, $\nu=\nu_c$, and $\omega=\omega_c$.
The chosen step size is admissible since $T\ge1$.

\emph{Step 3. Sum the expected descent inequalities.}
Take unconditional expectations in \eqref{eq:theory-one-step}, use
the tower property, and rearrange to get:
\[
\frac{\eta\kappa}{4}\mathbb E\|g_{c,t}^{\rm raw}\|^2
\le\mathbb EF(\boldsymbol{\phi}_t)-\mathbb EF(\boldsymbol{\phi}_{t+1})
+\frac{\eta\beta_c^2}{2\kappa}+\frac{L_c\eta^2\mathcal B}{2}.
\]
Summing from $t=0$ to $T-1$ telescopes the objective terms.
For a fixed initial state and $F\ge F_{\inf}$ this gives:
\[
\frac{\eta\kappa}{4}\sum_{t=0}^{T-1}\mathbb E\|g_{c,t}^{\rm raw}\|^2
\le\Delta_F+\frac{T\eta\beta_c^2}{2\kappa}
+\frac{TL_c\eta^2\mathcal B}{2}.
\]
Dividing by $\frac{\eta\kappa T}{4}$ gives the general constant-step bound:
\begin{equation}
\frac1{T}\sum_{t=0}^{T-1}\mathbb E\|g_{c,t}^{\rm raw}\|^2
\le\frac{4\Delta_F}{\eta\kappa T}
+\frac{2\beta_c^2}{\kappa^2}+\frac{2L_c\eta\mathcal B}{\kappa}.
\label{eq:theory-general-rate}
\end{equation}

Substituting $\eta=\frac{\kappa}{2L_c\mathcal A\sqrt{T}}$
into Eq.~\eqref{eq:theory-general-rate} proves Eq.~\eqref{eq:theory-rate}.
\end{proof}

For Gaussian RGE with $b_c$ independent coarse-space directions,
Theorem~\ref{prop:theory-rge} permits:
\[
\beta_c^2\le L_c^2\varepsilon^2d_cM_c^3,\quad
\nu_c=\frac{2(d_c+1)M_c^2}{b_c},\quad
\omega_c=\frac{L_c^2\varepsilon^2M_c^3}{2b_c}d_c(d_c+2)(d_c+4).
\]
Hence the ideal model has $O(T^{-\frac{1}{2}})+O(\varepsilon^2)$ stationarity,
with coarse-dimensional sampling constants.
Here $T$ is the number of optimization steps, and $S$ is the curvature
refresh interval used in the method.

We now extend the same descent argument to approximate corrections.
Their mean error changes the residual term, while their variance changes
the sampling constants in the stationarity bound.

\paragraph{Additional assumption for approximate correction.}
\begin{enumerate}[label=(A\arabic*),ref=A\arabic*,start=6,leftmargin=*]
\item \textbf{Controlled correction error.}\label{ass:theory-correction}
The predictable weight satisfies $0\le\lambda_t\le\lambda_{\max}\le1$,
and the correction estimate obeys, uniformly in $t$:
\[
\|\mathbb E_t\widehat\tau_t-\tau_t\|\le\beta_\tau,\qquad
\operatorname{Var}_t(\widehat\tau_t)
\le\nu_\tau\|g_{c,t}^{\rm raw}\|^2+\omega_\tau
\]
for nonnegative constants $\beta_\tau,\nu_\tau,\omega_\tau$.
\end{enumerate}

\begin{corollary}[Convergence with an approximate correction]
\label{cor:theory-approximate-rate}
Under Assumptions~\ref{ass:theory-smoothness} to \ref{ass:theory-correction},
update with $\widehat d_t=\widehat g_{c,t}+\lambda_t\widehat\tau_t$.
Define:
\begin{align*}
\beta_{\rm app}&=\beta_c+\lambda_{\max}\beta_\tau,\\
\nu_{\rm app}&=2\nu_c+2\lambda_{\max}^2\nu_\tau,\qquad
\omega_{\rm app}=2\omega_c+2\lambda_{\max}^2\omega_\tau,\\
\mathcal A_{\rm app}&=2M^2+\nu_{\rm app},\qquad
\mathcal B_{\rm app}=2\beta_{\rm app}^2+\omega_{\rm app}.
\end{align*}
For $\eta=\frac{\kappa}{2L_c\mathcal A_{\rm app}\sqrt{T}}$, the bound is:
\begin{equation}
\frac1{T}\sum_{t=0}^{T-1}\mathbb E\|g_{c,t}^{\rm raw}\|^2
\le\frac{8L_c\mathcal A_{\rm app}\Delta_F}{\kappa^2\sqrt{T}}
+\frac{\mathcal B_{\rm app}}{\mathcal A_{\rm app}\sqrt{T}}
+\frac{2\beta_{\rm app}^2}{\kappa^2}.
\label{eq:theory-approximate-rate}
\end{equation}
\end{corollary}
\begin{proof}
The reference direction remains
$d_t=(1-\lambda_t)g_{c,t}+\lambda_tg_{\star,t}$, so its alignment and
norm bounds are unchanged. The conditional mean error is:
\[
\mathbb E_t\widehat d_t-d_t
=(\mathbb E_t\widehat g_{c,t}-g_{c,t})
+\lambda_t(\mathbb E_t\widehat\tau_t-\tau_t).
\]
Its norm is at most $\beta_c+\lambda_{\max}\beta_\tau=\beta_{\rm app}$.
Let $\xi_{c,t}$ and $\xi_{\tau,t}$ be the two centered estimates,
and denote their trace variances by $V_{c,t}$ and $V_{\tau,t}$.
Then we have:
\begin{align*}
\operatorname{Var}_t(\widehat d_t)
&=V_{c,t}+\lambda_t^2V_{\tau,t}
+2\lambda_t\mathbb E_t\langle\xi_{c,t},\xi_{\tau,t}\rangle\\
&\le V_{c,t}+\lambda_t^2V_{\tau,t}
+2\lambda_t\sqrt{V_{c,t}V_{\tau,t}}\\
&\le2V_{c,t}+2\lambda_t^2V_{\tau,t}\\
&\le\nu_{\rm app}\|g_{c,t}^{\rm raw}\|^2+\omega_{\rm app}.
\end{align*}
The first inequality is conditional Cauchy-Schwarz, and the second
uses $2ab\le a^2+b^2$. Together they control the cross-covariance
for jointly sampled estimators.
Lemma~\ref{lem:theory-descent} now applies with the approximate constants.
Summing its inequality and substituting the new step size exactly as in
the preceding theorem gives \eqref{eq:theory-approximate-rate}.
\end{proof}

For a deterministic scalar-broadcast correction, the squared mean error
is the sum over the two branches of \eqref{eq:theory-scalar-extra} and
the conditional correction variance is zero. A uniform upper bound on
that sum therefore supplies $\beta_\tau^2$ in the corollary.
For random scalar estimates, their mean error and variance must instead
be bounded separately as in Proposition~\ref{prop:theory-noisy-mse}.
Thus the projection result supplies a correction-error bound, while
the descent lemma explains how this error enters the stationarity bound.

\section{Implementation Details}
\label{sec:app-vms-eo-details}

This appendix specifies the scalar correction used by \Engine\ and its
relation to the approximation error in Section~\ref{sec:theory-projection}.

\subsection{Scalar correction and update}
\label{sec:app-surrogate}

\paragraph{Branch notation.}
Boldface $\mathbf c_\ell,\mathbf c_{f,\ell}\in\mathbb R^{p_\ell}$ denotes the exact per-row
coefficients in Section~\ref{sec:theory-projection}, while $c_\ell,c_{f,\ell}$
denote scalar surrogates. For $\ell\in\{B,A\}$, define
$\mathbf X_B=\mathbf B$ and $\mathbf X_A=\mathbf A$.
For the B-branch, $p_B=m$, $\mathbf G_B=\mathbf G\mathbf A^\top$,
$\mathbf h_B=\mathbf G\mathbf v_f$, $\mathbf q_B=\mathbf v_{\!\mathbf B}$,
and $\mathbf s_B=\mathbf A\mathbf v_f$.
For the A-branch, $p_A=r$, $\mathbf G_A=\mathbf B^\top\mathbf G$,
$\mathbf h_A=\mathbf B^\top\mathbf G\mathbf v_f$,
$\mathbf q_A=\mathbf v_{\!\mathbf A}$, and $\mathbf s_A=\mathbf v_f$.
Here $\mathbf G=\nabla_{\!\mathbf W}f$ defines the reference contractions.
The matrices $\mathbf1_{p_\ell}\mathbf q_\ell^\top$ and
$\mathbf1_{p_\ell}\mathbf s_\ell^\top$ have the same shape as $\mathbf X_\ell$.

\paragraph{Scalar coefficients.}
The denominator-adjusted reference scalars are:
\begin{equation}
c_\ell=\frac{\mathbf1_{p_\ell}^\top\mathbf G_\ell\mathbf q_\ell}
{\rho+p_\ell\|\mathbf q_\ell\|^2},\qquad
c_{f,\ell}=\frac{\mathbf1_{p_\ell}^\top\mathbf h_\ell}
{\rho+p_\ell\|\mathbf v_f\|^2}.
\tag{D.1}\label{eq:app-scalar-surrogate}
\end{equation}
These scalars differ from the arithmetic row means
$\bar c_\ell$ and $\bar c_{f,\ell}$ in
Section~\ref{sec:theory-projection}.
For the exact preconditioned direction
$\mathbf g_{c,\ell}=\boldsymbol\Sigma_\ell\mathbf G_\ell$,
Eq.~\eqref{eq:app-sm-inverse} gives:
\begin{equation}
\langle\mathbf1_{p_\ell}\mathbf q_\ell^\top,\mathbf g_{c,\ell}\rangle_F
=\frac{\mathbf1_{p_\ell}^\top\mathbf G_\ell\mathbf q_\ell}
{\rho+\|\mathbf q_\ell\|^2}.
\tag{D.2}
\end{equation}
When $c_\ell\ne0$, the ratio of this contraction to $c_\ell$ is
$(\rho+p_\ell\|\mathbf q_\ell\|^2)/(\rho+\|\mathbf q_\ell\|^2)$,
which approaches $p_\ell$ when $\|\mathbf q_\ell\|^2\gg\rho$.
For the preconditioned RLOO estimate, we use:
\begin{equation}
c_\ell^{\mathrm{impl}}=
\frac{\langle\mathbf1_{p_\ell}\mathbf q_\ell^\top,
\widehat{\mathbf g}_{c,\ell}\rangle_F}
{\rho+p_\ell\|\mathbf q_\ell\|^2+10^{-8}}.
\tag{D.3}\label{eq:app-implementation-scalars}
\end{equation}
For a fine probe, let $\mathbf Z_k\in\mathbb R^{m\times n}$ be the
perturbation matrix for this weight block, $f_k$ the loss at the $k$-th
perturbed model, and $\bar f=K^{-1}\sum_{k=1}^K f_k$, with $K>1$.
Let $\mathbf v_{f,k}$ denote the curvature vector when the $k$-th
contribution is accumulated, before its next curvature update.
The sampled fine coefficients used in the correction are:
\begin{equation}
\begin{aligned}
c_{f,B}^{\mathrm{impl}}
&=\operatorname{EMA}_{\mathrm{bc}}\!\left[
\frac{1}{\varepsilon(K-1)}\sum_{k=1}^K
\frac{(f_k-\bar f)\mathbf1_m^\top\mathbf Z_k\mathbf v_{f,k}}
{\rho+m\|\mathbf v_{f,k}\|^2+10^{-8}}\right],\\
c_{f,A}^{\mathrm{impl}}
&=\operatorname{EMA}_{\mathrm{bc}}\!\left[
\frac{1}{\varepsilon(K-1)}\sum_{k=1}^K
\frac{(f_k-\bar f)\mathbf1_r^\top\mathbf B^\top\mathbf Z_k\mathbf v_{f,k}}
{\rho+r\|\mathbf v_{f,k}\|^2+10^{-8}}\right].
\end{aligned}
\tag{D.4}\label{eq:app-fine-implementation-scalars}
\end{equation}
Here $\operatorname{EMA}_{\mathrm{bc}}$ is an exponential moving average
with decay $0.9$, initialized at zero and divided by $1-0.9^j$
after $j$ fine probes in the current coarse phase.
Its state is reset at each new coarse phase and retained between probes.
The $10^{-8}$ term stabilizes the sampled-coefficient denominators.
The superscript $\mathrm{impl}$ identifies the coefficients used in the update,
including their estimation, normalization, and smoothing errors.

\paragraph{Correction and update.}
The scalar-corrected step, before clipping and momentum processing, is:
\begin{equation}
\begin{aligned}
\widehat{\boldsymbol\tau}^{\rm impl}_\ell
&=\mathbf1_{p_\ell}
(c_\ell^{\mathrm{impl}}\mathbf q_\ell^\top-c_{f,\ell}^{\mathrm{impl}}\mathbf s_\ell^\top),\\
\Delta\mathbf X_\ell
&=-\eta_\ell\bigl(\widehat{\mathbf g}_{c,\ell}
+\lambda_\ell\widehat{\boldsymbol\tau}^{\rm impl}_\ell\bigr).
\end{aligned}
\tag{D.5}\label{eq:app-deployed-update}
\end{equation}
With the current-basis directions defined above, this correction has the form
$\widetilde\tau_\ell$ in Proposition~\ref{prop:theory-projection} with
$\widetilde c_\ell=\rho c_\ell^{\mathrm{impl}}$ and
$\widetilde c_{f,\ell}=\rho c_{f,\ell}^{\mathrm{impl}}$.
These factors account for the $1/\rho$ in the exact residual.
Substituting them into Eq.~\eqref{eq:theory-scalar-extra} includes both
coefficient normalization and correction scaling in the error decomposition.

\paragraph{Initialization safeguards.}
When $\mathbf A$ is reset to zero, B-branch feedback uses
$\mathbf A_{\mathrm{old}}\mathbf v_f$ if a cached basis is available and
$\max_{i,j}|A_{ij}|<10^{-10}$.
Transferred curvature vectors are normalized only when their norms exceed
$10^{-10}$. Otherwise, their initialized values are retained.
The cached-basis fallback introduces a directional approximation in addition
to the scalar coefficient error analyzed above.

\section{Detailed Experiments and Results}
\label{sec:app-tables}

This appendix reports complete results and learning curves. Experimental settings and hyperparameters are collected in Appendix~\ref{sec:app-exp-setup}.

\subsection{Full Results and Ablation}
\label{sec:app-main-tables}
\label{sec:app-ablation-table}

Tables~\ref{tab:app-full-results} and~\ref{tab:app-ablation-full} expand the main results and ablation in Tables~\ref{tab:main-results} and~\ref{tab:ablation-vmseo}, reporting five-seed means and standard deviations. All results use accuracy (\%). Baselines are independently reproduced under the LOREN protocol. The ablation compares \Engine\ with the same V-cycle without VMS-EO correction. Bold and cyan bold mark the best and second-best results in each column. The highlighted rows report \Engine\ and its absolute accuracy gain over LOREN.

\begin{table}[!htbp]
\centering
\caption{Full results over five seeds, reported as mean accuracy (\%) $\pm$ standard deviation. Avg. averages the four task means.}
\label{tab:app-full-results}
\setlength{\tabcolsep}{4pt}
\renewcommand{\arraystretch}{1.05}
\resizebox{\linewidth}{!}{%
\small
\begin{tabular}{l ccccc ccccc}
\toprule
\textbf{GLUE} (FP32) & \multicolumn{5}{c}{\textbf{GPT-2-XL} (1.5B)} & \multicolumn{5}{c}{\textbf{OPT-2.7B}} \\
\cmidrule(lr){2-6}\cmidrule(lr){7-11}
Method & MNLI & QNLI & SST-2 & CoLA & Avg. & MNLI & QNLI & SST-2 & CoLA & Avg. \\
\midrule
MeZO        & 38.4\spm{1.0} & 59.2\spm{0.2} & 74.3\spm{1.0} & 65.8\spm{0.3} & 59.4 & 50.4\spm{3.3} & \snd{75.2\spm{1.0}} & \snd{88.7\spm{0.2}} & 68.4\spm{2.3} & 70.7 \\
MeZO-Adam   & \snd{50.8\spm{1.4}} & 71.9\spm{4.4} & \snd{91.2\spm{0.5}} & 71.4\spm{0.8} & 71.3 & 45.9\spm{1.2} & 72.3\spm{3.2} & 86.9\spm{2.8} & \snd{68.5\spm{1.0}} & 68.4 \\
MeZO-SVRG   & 49.8\spm{1.1} & 65.5\spm{1.0} & 84.5\spm{1.5} & 65.6\spm{0.4} & 66.4 & 41.5\spm{0.2} & 59.8\spm{1.0} & 62.9\spm{1.9} & 66.5\spm{0.2} & 57.7 \\
LOZO        & 42.9\spm{0.9} & 60.6\spm{1.1} & 80.0\spm{1.0} & 66.2\spm{0.5} & 62.4 & 43.2\spm{1.1} & 59.5\spm{0.8} & 66.2\spm{3.0} & 66.3\spm{0.3} & 58.8 \\
HiZOO       & 48.4\spm{0.2} & 66.8\spm{3.4} & 88.9\spm{0.2} & 71.2\spm{0.9} & 68.8 & 50.9\spm{1.0} & 74.7\spm{1.0} & 88.2\spm{0.6} & 66.3\spm{0.2} & 70.0 \\
LOREN       & 50.6\spm{0.5} & \snd{74.1\spm{1.2}} & 89.0\spm{0.9} & \snd{72.0\spm{0.8}} & \snd{71.4} & \snd{54.1\spm{0.7}} & 72.6\spm{1.8} & 88.2\spm{0.8} & 68.1\spm{0.2} & \snd{70.8} \\
\rowcolor{enginerow}
\textbf{\Engine} (Ours) & \best{52.0\spm{0.5}} & \best{77.0\spm{1.0}} & \best{91.8\spm{0.5}} & \best{72.7\spm{0.6}} & \best{73.4} & \best{55.5\spm{0.5}} & \best{75.4\spm{1.2}} & \best{91.0\spm{0.4}} & \best{69.5\spm{0.2}} & \best{72.9} \\
\rowcolor{enginerow}
$\Delta$ vs LOREN & \gain{1.4} & \gain{2.9} & \gain{2.8} & \gain{0.7} & \gain{2.0} & \gain{1.4} & \gain{2.8} & \gain{2.8} & \gain{1.4} & \gain{2.1} \\
\midrule
\textbf{SuperGLUE} (BF16) & \multicolumn{5}{c}{\textbf{LLaMA-3-8B}} & \multicolumn{5}{c}{\textbf{OPT-13B}} \\
\cmidrule(lr){2-6}\cmidrule(lr){7-11}
Method & RTE & BoolQ & WiC & CB & Avg. & RTE & BoolQ & WiC & CB & Avg. \\
\midrule
MeZO        & \snd{59.1\spm{1.5}} & 64.0\spm{0.8} & \snd{59.3\spm{1.9}} & 71.5\spm{1.5} & 63.5 & 59.0\spm{0.5} & 64.4\spm{1.1} & 57.1\spm{0.6} & 72.7\spm{1.5} & 63.3 \\
MeZO-Adam   & 58.7\spm{1.0} & 63.2\spm{0.2} & 57.1\spm{1.0} & 68.9\spm{0.8} & 62.0 & \snd{60.8\spm{1.5}} & 63.4\spm{1.3} & 55.7\spm{1.0} & 70.2\spm{1.7} & 62.2 \\
MeZO-SVRG   & 58.2\spm{1.5} & 63.1\spm{0.2} & 58.0\spm{1.1} & 64.9\spm{3.1} & 61.1 & 59.3\spm{1.0} & \best{65.2\spm{0.8}} & 56.8\spm{0.8} & 69.4\spm{3.0} & 62.7 \\
LOZO        & 57.8\spm{1.2} & 63.6\spm{0.7} & 57.2\spm{2.0} & 71.1\spm{4.2} & 62.4 & 58.2\spm{0.7} & 63.4\spm{0.2} & \snd{57.7\spm{1.6}} & 68.8\spm{3.0} & 62.0 \\
HiZOO       & 58.2\spm{0.5} & 63.6\spm{0.2} & 59.2\spm{1.6} & 67.1\spm{3.9} & 62.0 & 59.5\spm{1.2} & 63.7\spm{0.4} & 57.0\spm{1.1} & 72.7\spm{1.8} & 63.2 \\
LOREN       & 58.4\spm{1.2} & \snd{64.5\spm{1.3}} & 58.9\spm{1.3} & \snd{72.4\spm{2.7}} & \snd{63.6} & 59.5\spm{0.7} & 63.2\spm{0.6} & 57.0\spm{0.9} & \snd{73.7\spm{1.2}} & \snd{63.4} \\
\rowcolor{enginerow}
\textbf{\Engine} (Ours) & \best{59.8\spm{1.1}} & \best{66.0\spm{1.0}} & \best{59.8\spm{1.2}} & \best{73.2\spm{2.3}} & \best{64.7} & \best{62.1\spm{0.8}} & \snd{65.2\spm{0.6}} & \best{59.0\spm{0.8}} & \best{79.1\spm{1.2}} & \best{66.4} \\
\rowcolor{enginerow}
$\Delta$ vs LOREN & \gain{1.4} & \gain{1.5} & \gain{0.9} & \gain{0.8} & \gain{1.1} & \gain{2.6} & \gain{2.0} & \gain{2.0} & \gain{5.4} & \gain{3.0} \\
\bottomrule
\end{tabular}}
\end{table}

\begin{table}[!htbp]
\centering
\caption{VMS-EO ablation over five seeds, reported as mean accuracy (\%) $\pm$ standard deviation.}
\label{tab:app-ablation-full}
\setlength{\tabcolsep}{4pt}
\renewcommand{\arraystretch}{1.05}
\resizebox{\linewidth}{!}{%
\small
\begin{tabular}{l ccccc ccccc}
\toprule
\textbf{GLUE} (FP32) & \multicolumn{5}{c}{\textbf{GPT-2-XL} (1.5B)} & \multicolumn{5}{c}{\textbf{OPT-2.7B}} \\
\cmidrule(lr){2-6}\cmidrule(lr){7-11}
Method & MNLI & QNLI & SST-2 & CoLA & Avg. & MNLI & QNLI & SST-2 & CoLA & Avg. \\
\midrule
LOREN               & \snd{50.6\spm{0.5}} & \snd{74.1\spm{1.2}} & 89.0\spm{0.9} & \snd{72.0\spm{0.8}} & \snd{71.4} & 54.1\spm{0.7} & 72.6\spm{1.8} & 88.2\spm{0.8} & 68.1\spm{0.2} & 70.8 \\
\Engine\ w/o VMS-EO   & 48.8\spm{0.6} & 72.7\spm{1.1} & \snd{90.2\spm{0.5}} & 71.5\spm{0.7} & 70.8 & \snd{54.7\spm{0.8}} & \snd{73.8\spm{1.3}} & \snd{89.5\spm{0.4}} & \snd{69.1\spm{1.0}} & \snd{71.8} \\
\rowcolor{enginerow}
\textbf{\Engine} (Ours) & \best{52.0\spm{0.5}} & \best{77.0\spm{1.0}} & \best{91.8\spm{0.5}} & \best{72.7\spm{0.6}} & \best{73.4} & \best{55.5\spm{0.5}} & \best{75.4\spm{1.2}} & \best{91.0\spm{0.4}} & \best{69.5\spm{0.2}} & \best{72.9} \\
\rowcolor{enginerow}
$\Delta$ vs LOREN & \gain{1.4} & \gain{2.9} & \gain{2.8} & \gain{0.7} & \gain{2.0} & \gain{1.4} & \gain{2.8} & \gain{2.8} & \gain{1.4} & \gain{2.1} \\
\midrule
\textbf{SuperGLUE} (BF16) & \multicolumn{5}{c}{\textbf{LLaMA-3-8B}} & \multicolumn{5}{c}{\textbf{OPT-13B}} \\
\cmidrule(lr){2-6}\cmidrule(lr){7-11}
Method & RTE & BoolQ & WiC & CB & Avg. & RTE & BoolQ & WiC & CB & Avg. \\
\midrule
LOREN               & \snd{58.4\spm{1.2}} & 64.5\spm{1.3} & \snd{58.9\spm{1.3}} & \snd{72.4\spm{2.7}} & \snd{63.6} & \snd{59.5\spm{0.7}} & 63.2\spm{0.6} & 57.0\spm{0.9} & 73.7\spm{1.2} & 63.4 \\
\Engine\ w/o VMS-EO   & 58.2\spm{1.3} & \snd{65.2\spm{1.0}} & 57.8\spm{1.2} & 71.4\spm{2.5} & 63.2 & 59.0\spm{0.9} & \snd{64.1\spm{0.7}} & \snd{57.8\spm{0.9}} & \snd{75.0\spm{1.5}} & \snd{64.0} \\
\rowcolor{enginerow}
\textbf{\Engine} (Ours) & \best{59.8\spm{1.1}} & \best{66.0\spm{1.0}} & \best{59.8\spm{1.2}} & \best{73.2\spm{2.3}} & \best{64.7} & \best{62.1\spm{0.8}} & \best{65.2\spm{0.6}} & \best{59.0\spm{0.8}} & \best{79.1\spm{1.2}} & \best{66.4} \\
\rowcolor{enginerow}
$\Delta$ vs LOREN & \gain{1.4} & \gain{1.5} & \gain{0.9} & \gain{0.8} & \gain{1.1} & \gain{2.6} & \gain{2.0} & \gain{2.0} & \gain{5.4} & \gain{3.0} \\
\bottomrule
\end{tabular}}
\end{table}

\subsection{Additional Convergence Curves}
\label{sec:app-learning-curves}

\begin{figure}[!htbp]
  \centering
  \includegraphics[width=\linewidth]{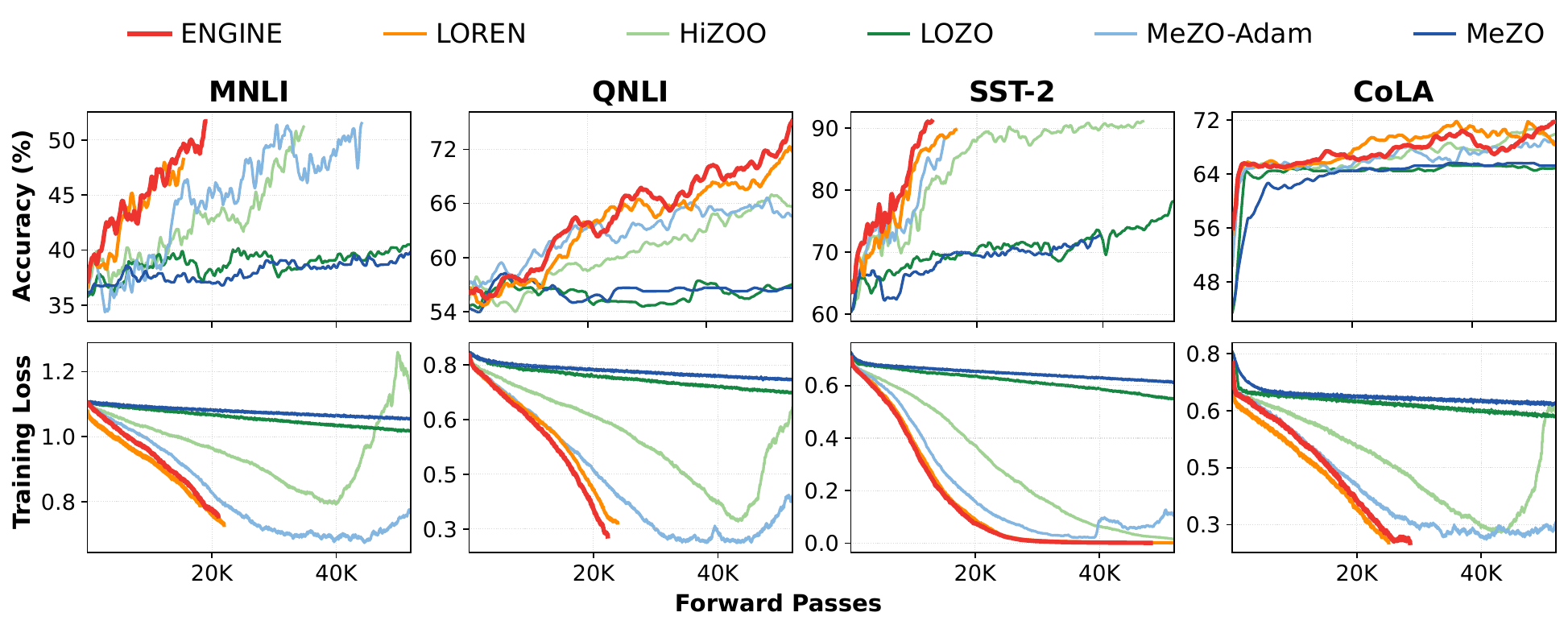}
  \caption{\textbf{GPT-2-XL (1.5B) training dynamics on four GLUE tasks (MNLI, QNLI, SST-2, CoLA).} \textit{Top row}: validation accuracy. \textit{Bottom row}: training loss.}
  \label{fig:app-curves-gpt2xl}
\end{figure}

\begin{figure}[!htbp]
  \centering
  \includegraphics[width=\linewidth]{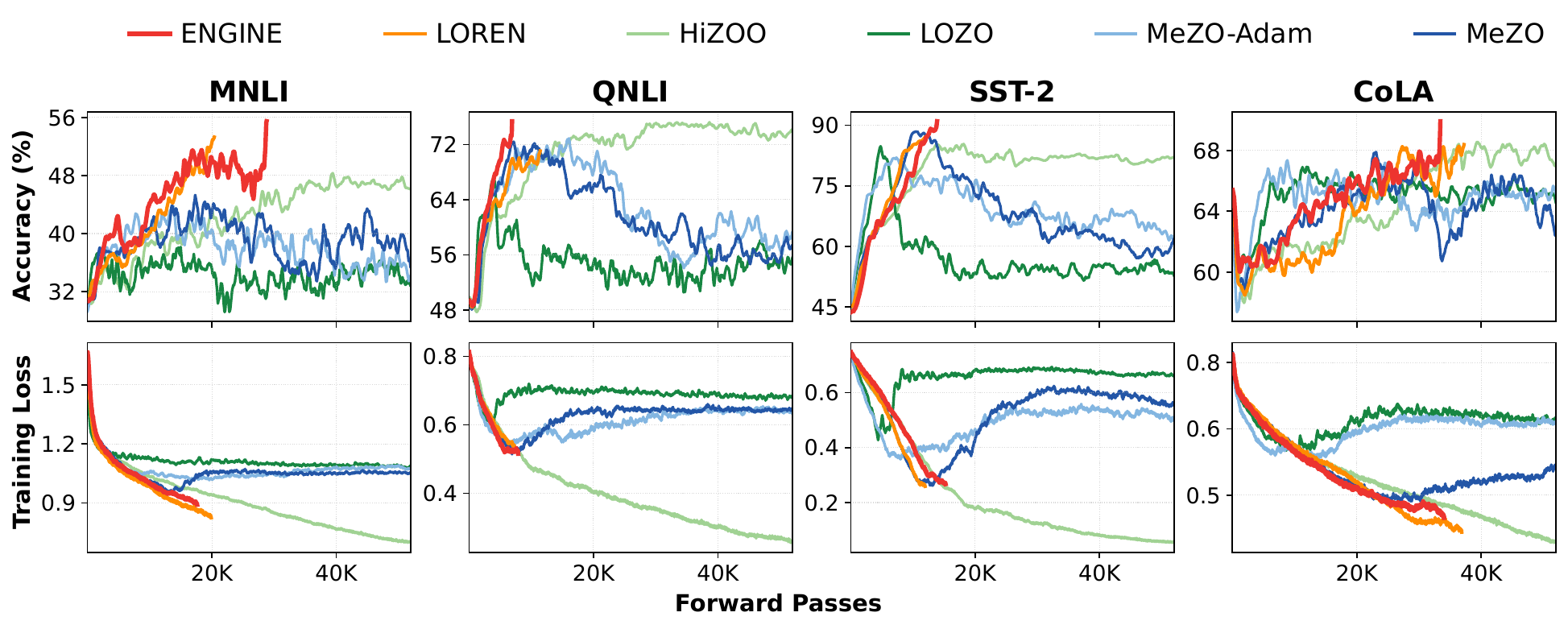}
  \caption{\textbf{OPT-2.7B training dynamics on four GLUE tasks (MNLI, QNLI, SST-2, CoLA).} \textit{Top row}: validation accuracy. \textit{Bottom row}: training loss.}
  \label{fig:app-curves-opt27b}
\end{figure}

\begin{figure}[!htbp]
  \centering
  \includegraphics[width=\linewidth]{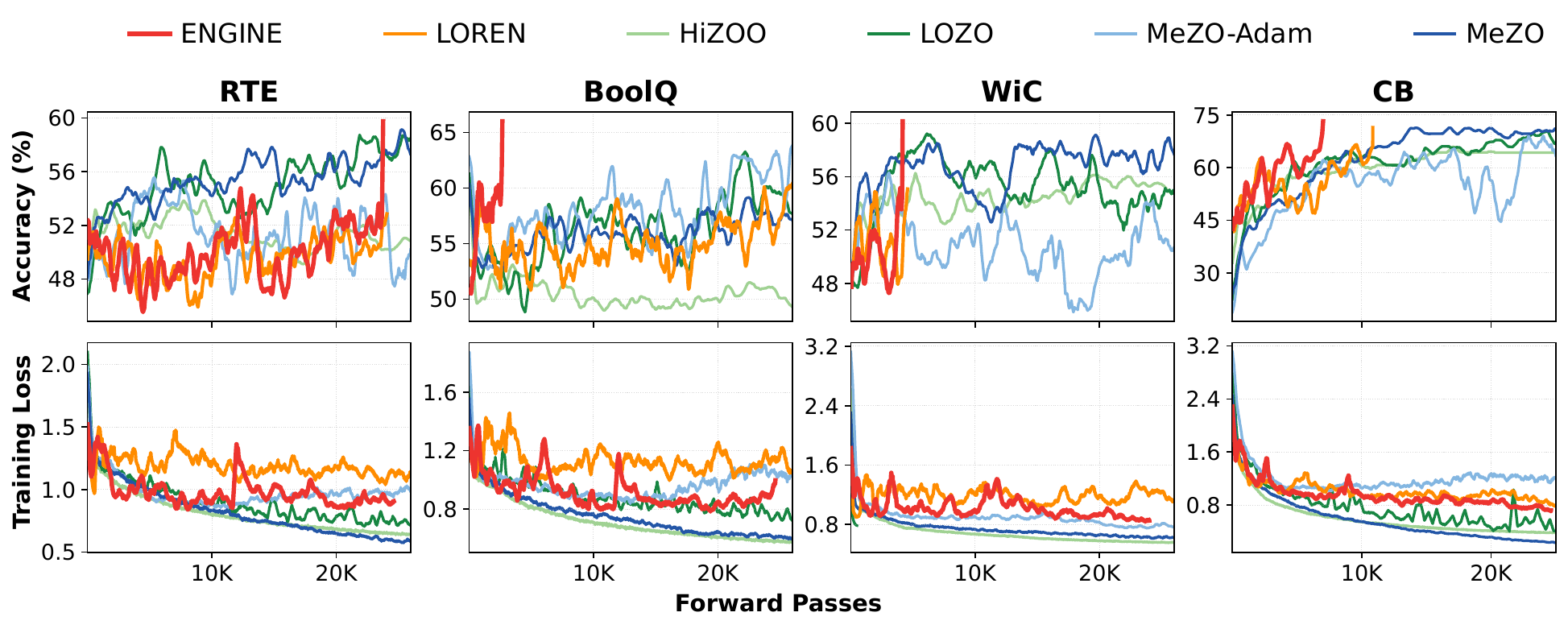}
  \caption{\textbf{LLaMA-3-8B training dynamics on four SuperGLUE tasks (RTE, BoolQ, WiC, CB).} \textit{Top row}: validation accuracy. \textit{Bottom row}: training loss.}
  \label{fig:app-curves-llama3-8b}
\end{figure}

\begin{figure}[!htbp]
  \centering
  \includegraphics[width=\linewidth]{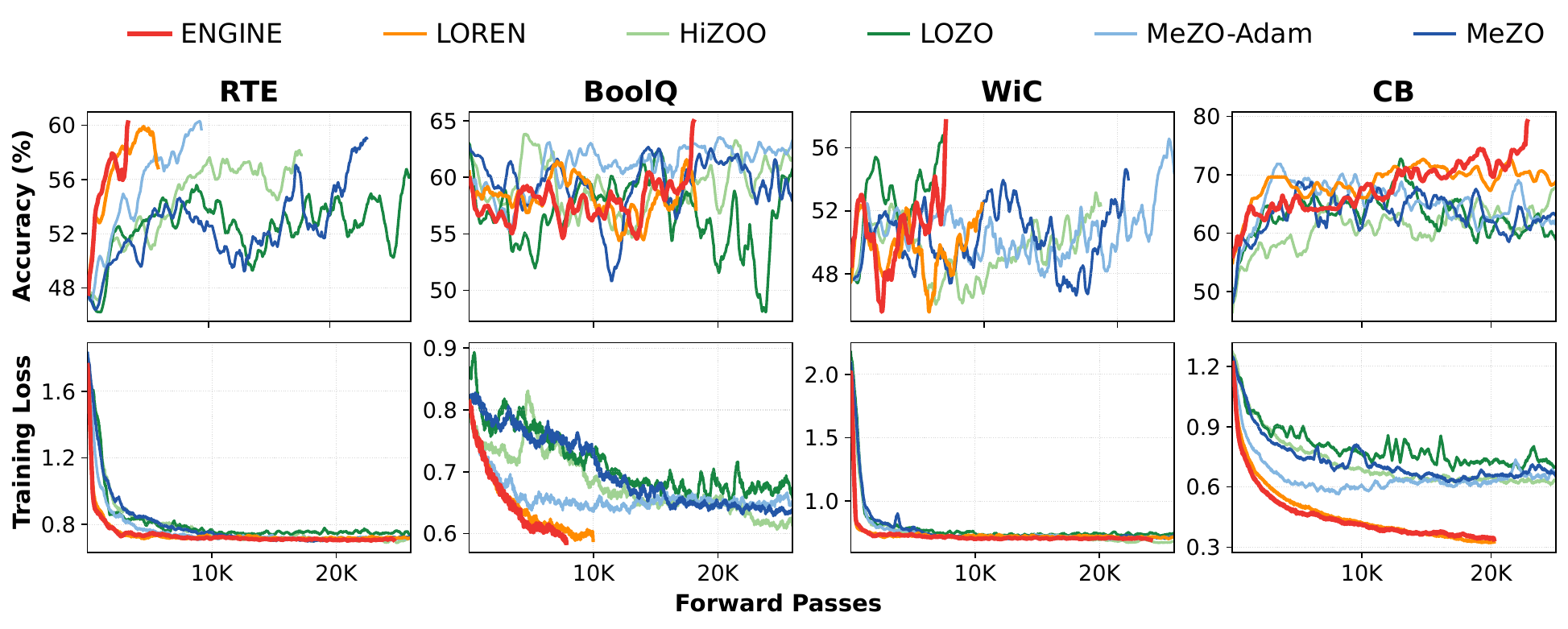}
  \caption{\textbf{OPT-13B training dynamics on four SuperGLUE tasks (RTE, BoolQ, WiC, CB).} \textit{Top row}: validation accuracy. \textit{Bottom row}: training loss.}
  \label{fig:app-curves-opt13b}
\end{figure}

\begin{figure}[!htbp]
  \centering
  \includegraphics[width=\linewidth]{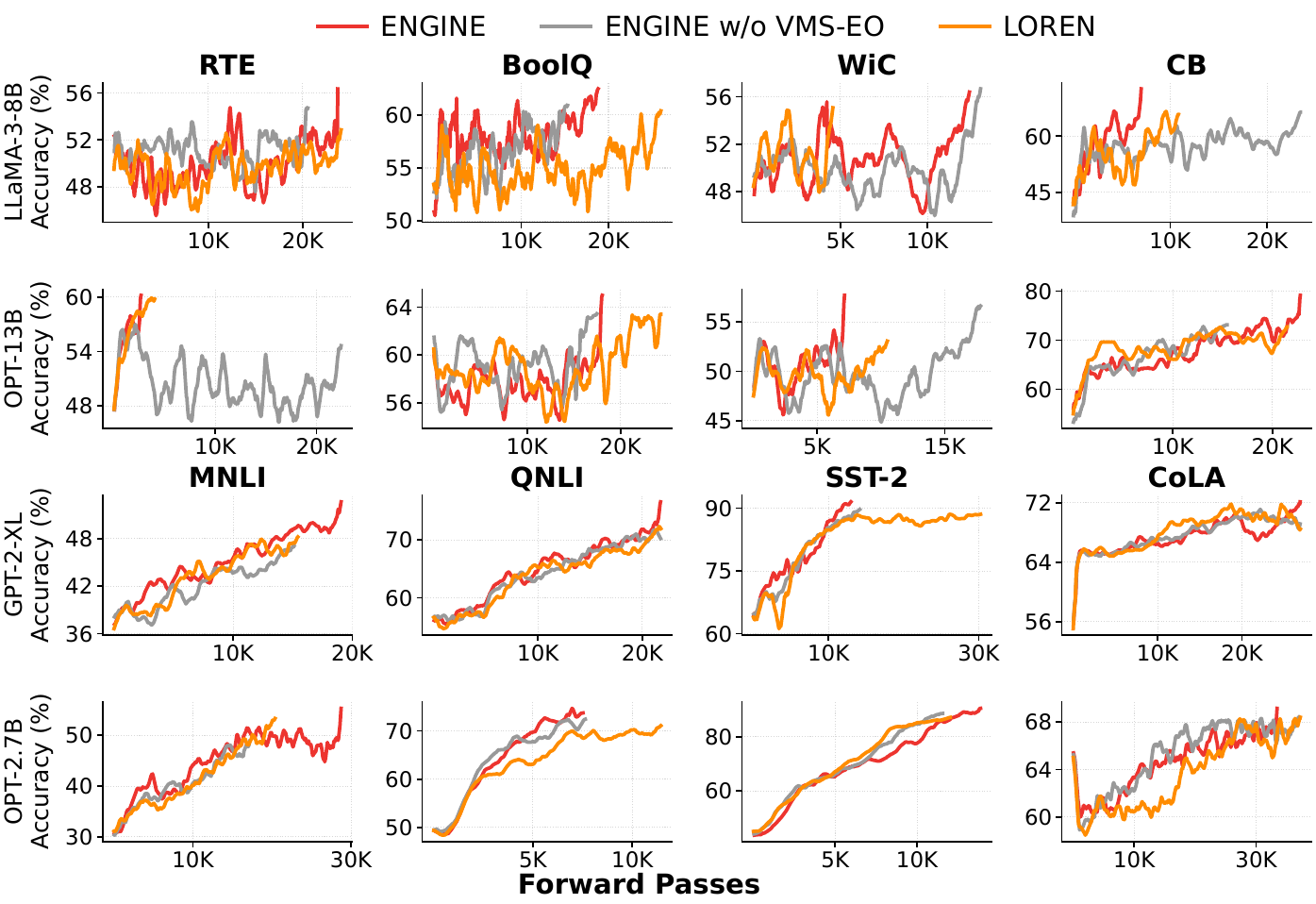}
  \caption{\textbf{VMS-EO ablation across four models and 16 tasks.}
  Validation accuracy versus forward passes for \Engine\ (red), \Engine\ w/o VMS-EO (grey), and LOREN (orange).
  From top to bottom, the rows show LLaMA-3-8B, OPT-13B, GPT-2-XL, and OPT-2.7B.
  In the first two rows, columns show RTE, BoolQ, WiC, and CB.
  In the last two rows, columns show MNLI, QNLI, SST-2, and CoLA.}
  \label{fig:app-ablation-4x4}
\end{figure}

Figures~\ref{fig:app-curves-gpt2xl} to \ref{fig:app-curves-opt13b} extend Figure~\ref{fig:exp-convergence} to all 16 model-task settings, reporting validation accuracy and training loss. Figure~\ref{fig:app-ablation-4x4} shows how the ablation results in Table~\ref{tab:app-ablation-full} evolve with the forward-pass budget.

\FloatBarrier

\section{Experimental Setup and Hyperparameters}
\label{sec:app-exp-setup}

\paragraph{Tasks and metrics.}
We follow the LOREN protocol for datasets, splits, and training. GPT-2-XL and OPT-2.7B use FP32 on MNLI, QNLI, SST-2, and CoLA. LLaMA-3-8B and OPT-13B use BF16 on RTE, BoolQ, WiC, and CB. All tasks use accuracy (\%) for the main results, ablation, learning curves, and sensitivity analyses. Avg. is the arithmetic mean of the four task accuracies for each model.

\paragraph{Training budget.}
Each run uses one NVIDIA A100 80GB GPU. All methods use batch size 64, perturbation radius $\epsilon=10^{-3}$, and direction count $K=6$. Training lasts 8,000 steps for GPT-2-XL and OPT-2.7B and 4,000 steps for LLaMA-3-8B and OPT-13B.

\paragraph{Method configurations.}
Across models, MeZO-Adam uses betas $(0.9,0.999)$. MeZO-SVRG uses full-batch update interval 2. LOZO searches ranks $\{\mathbf{2},4,8\}$ and intervals $\{\mathbf{50},100\}$. HiZOO uses Hessian smoothing $10^{-8}$. LOREN uses curvature learning rate $\nu=10^{-3}$ and damping $10^{-1}$.

Tables~\ref{tab:baseline-hyperparameters} and~\ref{tab:engine-hyperparameters} give the baseline and \Engine\ settings. Baseline settings follow LOREN, and bold entries mark selected values. The ablation uses the same \Engine\ settings with $\lambda=0$.

\begin{table}[!htbp]
\caption{Specific hyperparameters for the baselines. Learning rate search grids are listed by model, with selected values in bold. The two MeZO-SVRG rows give full-batch and mini-batch rates.}
\label{tab:baseline-hyperparameters}
\centering
\small
\setlength{\tabcolsep}{2.5pt}
\begin{tabular}{@{}lccc@{}}
\toprule
Method & GPT-2-XL & OPT-2.7B & LLaMA-3-8B, OPT-13B \\
\midrule
MeZO & \{1e-4,\,5e-5,\,1e-5,\,\textbf{5e-6},\,1e-6\} & \{1e-4,\,5e-5,\,1e-5,\,\textbf{5e-6},\,1e-6\} & \{1e-5,\,5e-6,\,\textbf{1e-6},\,5e-7,\,1e-7\} \\
MeZO-Adam & \{1e-3,\,5e-4,\,\textbf{1e-4},\,5e-5,\,1e-5\} & \{1e-3,\,5e-4,\,1e-4,\,\textbf{5e-5},\,1e-5\} & \{1e-3,\,5e-4,\,1e-4,\,\textbf{5e-5},\,1e-5\} \\
MeZO-SVRG full & \{1e-4,\,\textbf{5e-5},\,1e-5,\,5e-6,\,1e-6\} & \{1e-4,\,\textbf{5e-5},\,1e-5,\,5e-6,\,1e-6\} & \{1e-5,\,5e-6,\,\textbf{1e-6},\,5e-7,\,1e-7\} \\
MeZO-SVRG mini & \{1e-5,\,5e-6,\,\textbf{1e-6},\,5e-7,\,1e-7\} & \{1e-5,\,5e-6,\,\textbf{1e-6},\,5e-7,\,1e-7\} & \{1e-6,\,5e-7,\,\textbf{1e-7},\,5e-8,\,1e-8\} \\
LOZO & \{1e-4,\,5e-5,\,1e-5,\,\textbf{5e-6},\,1e-6\} & \{1e-4,\,5e-5,\,1e-5,\,\textbf{5e-6},\,1e-6\} & \{1e-5,\,5e-6,\,1e-6,\,\textbf{5e-7},\,1e-7\} \\
HiZOO & \{1e-3,\,5e-4,\,1e-4,\,\textbf{5e-5},\,1e-5\} & \{1e-4,\,5e-5,\,1e-5,\,\textbf{5e-6},\,1e-6\} & \{1e-5,\,5e-6,\,\textbf{1e-6},\,5e-7,\,1e-7\} \\
LOREN & \{\textbf{1e-5},\,5e-6,\,1e-6,\,5e-7,\,1e-7\} & \{1e-5,\,5e-6,\,1e-6,\,\textbf{5e-7},\,1e-7\} & \{1e-5,\,5e-6,\,1e-6,\,\textbf{5e-7},\,1e-7\} \\
\Engine & \{\textbf{1e-5},\,5e-6,\,1e-6,\,5e-7,\,1e-7\} & \{1e-5,\,5e-6,\,1e-6,\,\textbf{5e-7},\,1e-7\} & \{1e-5,\,5e-6,\,1e-6,\,\textbf{5e-7},\,1e-7\} \\
\bottomrule
\end{tabular}

\bigskip
\normalsize
\caption{Specific hyperparameters for \Engine. Each row gives a model's learning rate, rank, and refresh intervals $S$ in coarse steps. All four models use $\nu=10^{-3}$, $\rho=10^{-3}$, C60F60, and mixing bounds $[0.3,0.8]$.}
\label{tab:engine-hyperparameters}
\setlength{\tabcolsep}{5pt}
\begin{tabular}{@{}lcccccc@{}}
\toprule
 & & & \multicolumn{4}{c}{Refresh interval $S$} \\
\cmidrule(l){4-7}
Model & $\eta$ & $r$ & MNLI & QNLI & CoLA & SST-2 \\
\midrule
GPT-2-XL & 1e-5 & 16 & 5 & 20 & 20 & 20 \\
OPT-2.7B & 1e-5 & 32 & 20 & 20 & 20 & 5 \\
\midrule
Model & $\eta$ & $r$ & CB & RTE & BoolQ & WiC \\
\midrule
LLaMA-3-8B & 5e-7 & 32 & 20 & 20 & 20 & 20 \\
OPT-13B & 5e-7 & 64 & 20 & 20 & 10 & 100 \\
\bottomrule
\end{tabular}
\end{table}

\FloatBarrier

\subsection{Hyperparameter Sensitivity}
\label{sec:app-hyperparameter-sensitivity}

Let $\mathcal I=\{5,10,20,100,1000\}$ be the candidate refresh intervals. The index $j$ identifies one of the 16 model and task settings, and $a_j(S)$ denotes its raw validation accuracy at refresh interval $S$. For a positive accuracy range across the candidates, we define the normalized score:
\begin{equation}
q_j(S)=\frac{a_j(S)-\min_{u\in\mathcal I}a_j(u)}
{\max_{u\in\mathcal I}a_j(u)-\min_{u\in\mathcal I}a_j(u)},
\qquad S\in\mathcal I.
\label{eq:refresh-normalized-score}
\end{equation}
Table~\ref{tab:refresh-sensitivity} reports these per-setting Min-Max scores. At $S=20$, they range from 0.91 to 1.00 across all 16 settings, supporting a common default of 20 coarse steps. The per-task intervals used for the main results are listed in Table~\ref{tab:engine-hyperparameters}.

\begin{table}[!htbp]
\caption{Sensitivity to the refresh interval $S$ in coarse steps. Entries are the normalized scores $q_j(S)$ in Eq.~\eqref{eq:refresh-normalized-score}. Bold marks the highest score for each model and task.}
\label{tab:refresh-sensitivity}
\centering
\normalsize
\setlength{\tabcolsep}{4pt}
\begin{tabular}{@{}rcccc@{\hspace{1.5em}}rcccc@{}}
\toprule
\multicolumn{5}{c}{GPT-2-XL} & \multicolumn{5}{c}{OPT-2.7B} \\
\cmidrule(lr){1-5}\cmidrule(l){6-10}
$S$ & MNLI & QNLI & CoLA & SST-2 & $S$ & MNLI & QNLI & CoLA & SST-2 \\
\midrule
5 & \textbf{1.00} & \textbf{1.00} & 0.94 & \textbf{1.00} & 5 & \textbf{1.00} & \textbf{1.00} & \textbf{1.00} & \textbf{1.00} \\
10 & 0.95 & \textbf{1.00} & \textbf{1.00} & 0.94 & 10 & \textbf{1.00} & 0.92 & 0.94 & 0.91 \\
20 & 0.92 & \textbf{1.00} & \textbf{1.00} & \textbf{1.00} & 20 & \textbf{1.00} & \textbf{1.00} & \textbf{1.00} & 0.91 \\
100 & 0.34 & 0.27 & 0.35 & 0.38 & 100 & 0.31 & 0.36 & 0.45 & 0.22 \\
1000 & 0.00 & 0.00 & 0.00 & 0.00 & 1000 & 0.00 & 0.00 & 0.00 & 0.00 \\
\midrule
\multicolumn{5}{c}{LLaMA-3-8B} & \multicolumn{5}{c}{OPT-13B} \\
\cmidrule(lr){1-5}\cmidrule(l){6-10}
$S$ & RTE & BoolQ & WiC & CB & $S$ & RTE & BoolQ & WiC & CB \\
\midrule
5 & \textbf{1.00} & \textbf{1.00} & 0.96 & \textbf{1.00} & 5 & \textbf{1.00} & \textbf{1.00} & \textbf{1.00} & \textbf{1.00} \\
10 & 0.96 & \textbf{1.00} & 0.93 & 0.91 & 10 & 0.92 & \textbf{1.00} & \textbf{1.00} & \textbf{1.00} \\
20 & \textbf{1.00} & \textbf{1.00} & \textbf{1.00} & \textbf{1.00} & 20 & \textbf{1.00} & 0.92 & 0.93 & \textbf{1.00} \\
100 & 0.41 & 0.53 & 0.61 & 0.57 & 100 & 0.61 & 0.67 & \textbf{1.00} & 0.77 \\
1000 & 0.00 & 0.00 & 0.00 & 0.00 & 1000 & 0.00 & 0.00 & 0.00 & 0.00 \\
\bottomrule
\end{tabular}
\end{table}

We also vary the coarse and fine phase lengths $M,N\in\{10,30,60\}$ epochs, following Algorithm~\ref{alg:engine-main}. In this schedule, each epoch comprises eight minibatch updates, so C60F60 allocates 480 coarse and 480 fine updates per cycle. For each schedule, we average validation accuracy over the 16 model and task settings, then apply Min-Max normalization across the nine schedules. Table~\ref{tab:schedule-sensitivity} shows that C60F60 achieves the highest average validation accuracy.

\begin{table}[!htbp]
\caption{Sensitivity to the V-cycle schedule. Rows give coarse phase length $M$ and columns give fine phase length $N$, both in epochs. Scores are Min-Max normalized averages over 16 model and task settings. Bold marks the best schedule.}
\label{tab:schedule-sensitivity}
\centering
\normalsize
\begin{tabular}{@{}lccc@{}}
\toprule
 & $N=60$ & $N=30$ & $N=10$ \\
\midrule
$M=60$ & \textbf{1.00} & 0.65 & 0.44 \\
$M=30$ & 0.91 & 0.76 & 0.34 \\
$M=10$ & 0.82 & 0.71 & 0.00 \\
\bottomrule
\end{tabular}
\end{table}

\FloatBarrier

\section{Efficiency Analysis}
\label{sec:app-system}

In Table~\ref{tab:common-target-efficiency}, \Engine\ has the highest query efficiency in all four settings and the highest wall-clock efficiency on GPT-2-XL and OPT-2.7B.

\begin{table}[!htbp]
\caption{Query and wall-clock speed-ups at common target accuracies. Each entry divides the largest query count or time in its column by that method's value, assigning the least efficient method $1.0\times$. Higher is better, and bold marks the best result.}
\label{tab:common-target-efficiency}
\centering
\normalsize
\setlength{\tabcolsep}{4pt}
\begin{tabular}{@{}l*{8}{c}@{}}
\toprule
Method & \multicolumn{2}{c}{GPT-2-XL} & \multicolumn{2}{c}{OPT-2.7B} & \multicolumn{2}{c}{LLaMA-3-8B} & \multicolumn{2}{c}{OPT-13B} \\
 & \multicolumn{2}{c}{SST-2 (70\%)} & \multicolumn{2}{c}{QNLI (70\%)} & \multicolumn{2}{c}{CB (65\%)} & \multicolumn{2}{c}{CB (65\%)} \\
\cmidrule(lr){2-3}\cmidrule(lr){4-5}\cmidrule(lr){6-7}\cmidrule(l){8-9}
 & Queries & Time & Queries & Time & Queries & Time & Queries & Time \\
\midrule
MeZO & 1.0 & 1.0 & 1.0 & 1.0 & 1.2 & 1.2 & 2.3 & 2.5 \\
MeZO-Adam & 10.3 & 9.8 & 9.7 & 9.2 & 1.0 & 1.0 & 1.0 & 1.0 \\
MeZO-SVRG & 3.9 & 2.1 & 4.1 & 4.5 & 2.1 & 1.8 & 3.2 & 2.9 \\
LOZO & 1.6 & 4.7 & 2.2 & 6.2 & 2.2 & \textbf{6.4} & 3.5 & \textbf{7.3} \\
HiZOO & 7.5 & 3.2 & 8.0 & 3.7 & 1.6 & 1.3 & 1.9 & 1.5 \\
LOREN & 12.7 & 9.5 & 11.7 & 8.3 & 4.6 & 2.0 & 5.0 & 3.4 \\
\midrule
\rowcolor{enginerow}
\Engine & \textbf{16.5} & \textbf{11.4} & \textbf{22.2} & \textbf{16.7} & \textbf{10.0} & 4.1 & \textbf{5.3} & 4.0 \\
\bottomrule
\end{tabular}
\end{table}

\section{Limitations and Future Work}
\label{sec:app-limitations}

\paragraph{Adaptive query allocation.}
The refresh interval $S$ and coarse/fine phase lengths are fixed in advance.
Curvature drift can make cached fine-space information stale, while the
relative benefit of each phase can change during training. Adapting refreshes
and phase lengths to curvature drift or validation-loss plateaus could improve
query allocation.

\paragraph{Learning-rate selection.}
The branch learning rates $\eta_{\!\mathbf B}$, $\eta_{\!\mathbf A}$, and
$\eta_f$ share one value to limit tuning, leaving separate rates to be explored.

\paragraph{Access requirements and task coverage.}
\Engine\ requires access to model parameters to perturb weights and transfer
states between spaces, limiting its use through prediction-only APIs.
Our experiments cover classification fine-tuning across four models and
16 model-task settings. Evaluation on generation tasks would establish
whether the observed benefits extend to other fine-tuning objectives.

\FloatBarrier

\end{document}